# Analyzing and Controlling Diversity in Quantum-Behaved Particle Swarm Optimization

**Li-Wei Li**  7191905002@stu.jiangnan.edu.cn

School of IoT Engineering, Jiangnan University, No 1800, Lihu Avenue, Wuxi, Jiangsu 214122

**Jun Sun**  sunjun_wx@hotmail.com

School of Artificial Intelligence and Computer Science, Jiangnan University, No 1800, Lihu Avenue, Wuxi, Jiangsu 214122

**Chao Li**  lcmeteor@hotmail.com

School of Artificial Intelligence and Computer Science, Jiangnan University, No 1800, Lihu Avenue, Wuxi, Jiangsu 214122

**Wei Fang**  wxfangwei@hotmail.com

School of Artificial Intelligence and Computer Science, Jiangnan University, No 1800, Lihu Avenue, Wuxi, Jiangsu 214122

**Vasile Palade**  ab5839@coventry.ac.uk

Center for Computational Science and Mathematical Modeling, Coventry University, Priory Street, Coventry, CV1 5FB, UK

**Xiao-Jun Wu**  wu_xiaojun@jiangnan.edu.cn

School of Artificial Intelligence and Computer Science, Jiangnan University, No 1800, Lihu Avenue, Wuxi, Jiangsu 214122

**Abstract:** This paper addresses the issues of controlling and analyzing the population diversity in quantum-behaved particle swarm optimization (QPSO), which is an optimization approach motivated by concepts in quantum mechanics and PSO. In order to gain an in-depth understanding of the role the diversity plays in the evolving process, we first define the genotype diversity by the distance to the average point of the particles' positions and the phenotype diversity by the fitness values for the QPSO. Then, the correlations between the two types of diversities and the search performance are tested and analyzed on several benchmark functions, and the distance-to-average-point diversity is showed to have stronger association with the search performance during the evolving processes. Finally, in the light of the performed diversity analyses, two strategies for controlling the distance-to-average-point diversities are proposed for the purpose of improving the search ability of the QPSO algorithm. Empirical studies on the QPSO with the introduced diversity control methods are performed on a set of benchmark functions from the CEC 2005 benchmark suite. The performance of the proposed methods are evaluated and compared with the original QPSO and other PSO variants.

**Keywords:** Swarm diversity, particle swarm optimization, diversity control, quantum behavior

## 1 Introduction

The particle swarm optimisation (PSO) algorithm falls into the category of swarm intelligence algorithms and is a population-based optimisation technique originally developed by Kennedy and Eberhart (1995). It was motivated by the social behaviour (i.e., collective behaviour) of bird flocking or fish schooling and shares many similarities with evolutionary computation techniques, such as genetic algorithms (GAs). The optimization process of a PSO algorithm begins with an initial population of random solutions and it then searches for optima by updating the individuals at each iteration. However, unlike GAs, PSO does not have evolution operators such as crossover and mutation. The potential solutions are known as particles, which fly through the solution space by following their own experiences and the current best particles.

During the last decade, the PSO has gained increasing popularity due to its effectiveness in performing difficult optimization tasks. It has been successfully applied in many research and application areas (Poli, 2007, 2008a). Some researchers have made theoretical analyses in order to gain deep insights into the search mechanism of PSO (Kennedy, 1998; Clerc and Kennedy, 2002; Eberhart and Shi, 1998; Trelea, 2003; Emara and Fattah, 2004; Gavi and Passino, 2003; Kadirkamanathan et al., 2006; Jiang et al., 2007; Poli, 2008b).

Most of these analyses were focused on the trajectory stability of the individual particle, which is the key to the understanding of the search mechanism and parameter selection of the algorithm. For example, Clerc and Kennedy undertook the first formal analysis of the particle trajectory and its stability properties (Clerc and Kennedy, 2002).

A considerable amount of work has been done in developing the original version of PSO. Shi and Eberhart (1998a) introduced the concept of inertia weight to the original version of PSO, in order to balance the local and global search during the optimization process. Clerc (1999) proposed an alternative version of PSO, by incorporating a parameter called constriction factor which should replace the restriction on the velocity. These two versions of PSO, known as the PSO with inertia weight (PSO-In) and the PSO with constriction factor (PSO-Co), have laid the foundation for further enhancement of PSO. In the original PSO, PSO-In and PSO-Co, the search of the particles is guided by the global best position and their personal best positions. This neighborhood topology is known as the global best model. Many researchers have also turn to the investigation of other neighborhood topologies, known as the local best models, first studied by Eberhart and Kennedy (1995) and subsequently, in depth, by many other researchers (Suganthan, 1999; Kennedy, 1999; Kennedy, 2002; Mendes et al., 2004; Liang and Suganthan, et al., 2005; Mohais et al., 2005). The objective there was to find other possible topologies to improve the performance of PSO. Generally speaking, the common purpose of the work mentioned above was to make a better balance between the exploration and the exploitation of the evolving particle swarm.

In PSO, the particles essentially follow a deterministic trajectory defined by a velocity update formula with two random acceleration coefficients. This is a semi-deterministic search, which restricts the search scope of each particle and may weaken the global search ability of the algorithm, particularly, at the later stage of search process. In view of this limitation, several probabilistic PSO algorithms simulate the particle trajectories by direct sampling, using a random number generator, from a distribution of practical interests (Secrest and Lamont, 2003; Richer and Blackwell, 2006; Kennedy, 2003; Krohling and Coelho, 2006; Sun et al., 2004a). The bare bones PSO (BBPSO) family are typical probabilistic PSO algorithms (Kennedy, 2003, 2004, 2006). In BBPSO, each particle does not have a velocity vector, but with its new position being sampled "around" a supposedly good one, according to a certain probability distribution, such as the Gaussian distribution (Kennedy, 2003).

The quantum-behaved particle swarm optimization (QPSO) algorithm, which is the focus of this paper, is also a probabilistic PSO motivated by the quantum mechanics and the trajectory analysis of PSO (Sun et al., 2004a, 2004b, 2005). The QPSO uses a strategy based on a quantum $\delta$ potential well to sample new positions directly around the previous best points, and its iterative equation is very different from that of the original PSO (Sun et al. 2004a). Besides, the QPSO needs no velocity vectors for particles and, essentially, belongs to the BBPSO family, but it samples the new position with a double exponential distribution. In addition, its update equation uses an adaptive strategy and has fewer parameters to be adjusted, leading to a good algorithmic performance as an overall result.

Empirical evidence has shown that the QPSO algorithm works well and has been successfully used to solve a wide range of continuous optimization problems. Many efficient strategies have been proposed to improve the performance of the algorithm (Liu et al., 2005; Wang and Zhou, 2007; Coelho, 2008; Huang et al., 2009; Pant et al., 2009; Sun et al., 2012a). A recent extensive survey of the QPSO and its application areas can be found in (Fang et al., 2010; Sun et al., 2011).

In order to gain a deep insight into how the QPSO works, in (Sun, et al., 2012b), we made a comprehensive theoretical analysis of the stochastic dynamical behaviour of the individual particle in QPSO, in terms of probability measure, derived the sufficient and necessary condition for the particle to be convergent or probabilistically bounded, and thus provided the guidelines for parameter selection of the algorithm based on these theoretical results. The goal of this paper is to make comprehensive analyses of the diversity of the whole particle swarm of the QPSO, and propose some diversity control strategies in order to improve the performance of the algorithm.

First, two types of diversity measures, namely, the distance-to-average-point and the entropy diversity, are defined for the QPSO algorithm, and then the correlation between diversity and fitness is tested on several benchmark problems in order to identify the roles that the two different diversity measures play in the evolving process. Based on the diversity analysis, two methods for controlling the distance-to-average-point diversity of the QPSO are proposed in order to improve the performance of the algorithm. The approaches are tested on the first fourteen benchmark functions from the CEC2005 benchmark suite, and their performance are compared with the original QPSO and other PSO variants in order to show the effectiveness of the proposed methods.

The rest of the paper is organized as follows. In section 2, we survey the related work on diversity issues within the evolutionary algorithms area. Section 3 provides a brief introduction of the QPSO algorithm. Section 4 analyses the diversity of the QPSO, and section 5 proposes two diversity control strategies for the QPSO algorithm. In section 6, the experimental results are provided on the benchmark functions. Some concluding remarks are given in the last section.

## 2. Related Work

As our attention mainly focuses on the diversity analysis and control for QPSO, in this section we make a survey of the related work within the evolutionary algorithms area with respect to the following three aspects.

### 2.1. Diversity Measures

Biological diversity refers to the difference between individuals in a population, which, by the nature of biology, implies a structural and behavioral difference. In evolutionary algorithms, structural difference refers to the fact that two individuals are not identical, while behavioral difference implies that the fitness values of two individuals are different. Thus, generally, there are two types of diversity. One is the genotype diversity that symbolizes the variety of individual structures, and the other is the phenotype diversity, which measures the behavioral difference of the population (Morrison and De Jong, 2001).

Barker and Martin (2000) proposed a distance-based population diversity measure for evolutionary algorithms. Kim et al (2003) introduced the concept of population diversity, to measure the diversity in a binary genetic algorithm, based on the Hamming distance between two individuals. Qi and Palmieri (1994a) measured the diversity of population in genetic algorithms by the multi-dimensional density of population. Ursem (2002) defined the diversity of population in a real-coded evolutionary algorithm by a distance-to-average-point measure. Koza (1992) used the term *variety* to denote the number of different genotypes contained in a population in genetic programming. McPhee and Hopper (1999) introduced several techniques for measuring the diversity of a population based on the genetic history of the individuals of genetic programming, and applied these measures to the genetic histories of several runs of four different problems. A more detailed survey of diversity measures in genetic programming can be found in (Burke et al., 2004)

The diversity measures mentioned above are all genotype diversities. The measure of success in evolutionary algorithms is typically the fitness of a solution or behavior in the problem's environment. Measures based on behavior compare differences among the populations' fitness values at a given time. Goldberg and Rudnick (1991) used variance of fitness to measure the diversity of the population. Rosca (1995) used the fitness values in a population to define an entropy and free energy measure. Entropy represents the amount of disorder of the population, where an increase in entropy represents an increase in diversity. Rosca's experiments showed that populations in evolutionary algorithms appeared to be stuck in local optima when entropy did not change or decrease monotonically in successive generations. By extending Rosca's work, Liu et al. (2007) proposed other three kinds of entropy - linear entropy, Gaussian entropy, and fitness proportional entropy – in order to express the diversity of an evolutionary algorithm.

### 2.2 Diversity Analysis

Some researchers have studied diversity issues to find unusual behavior of the evolving populations of evolutionary algorithms. It was showed that premature convergence of genetic algorithms was generally the result of rapid decline of population diversity (Booker, 1987). Barker and Martin (2000) defined the sum over all pairwise population distances as a measure of the population diversity genetic algorithms and investigated the time evolution of the expected diversity of a population. Goldberg and Rudnick applied variance-of-fitness diversity to two important problems in genetic algorithm theory, i.e., population sizing and the calculation of rigorous probabilistic convergence bounds (Goldberg and Rudnick, 1991). Qi and Palmieri discussed the unique diversification role of the crossover operator in genetic algorithms by using population density as diversity measure of the population (Qi and Palmeiri, 1994b). Friedrich et al. compared some well-known diversity mechanisms for evolutionary algorithms like deterministic crowding, fitness sharing, and others, with an algorithm without diversification, showing that diversification was necessary for global exploration (Friedrich et al., 2008). O'Reily (1997) discussed the dynamics of structural distance measures of the genetic programming population. Burke et al. found, through experiments, the varying correlation between diversity and fitness during different stages of the evolutionary process of genetic programming (Burke et al., 2004). They showed that populations in the genetic programming algorithm become structurally similar while maintaining a high amount of behavioral differences.

### 2.3 Diversity Promotion and Controlling

The canonical view of evolution and diversity is that high diversity provides more opportunities for evolution. However, as noted in several diversity studies, typical evolutionary algorithms contain a phase of exploration followed by exploitation (Ekárt and Németh, 2002). Promoting or preserving all kinds of diversity during the entire evolutionary process could be counterproductive to the exploitation phase. The type and amount of diversity required at different evolutionary times remains rather unclear. Nevertheless, several measures and approaches have been used to promote diversity. These strategies include crowding (DeJong, 1975; Mahfoud, 1992), preselection (Mahfoud, 1992), neighborhood (Collins, 1992), islands model (Martin et al., 2000), fitness sharing (Goldberg and Richardson, 1987), genotype sharing (Deb and Goldberg, 1989), the trigged hypermutation method (Cobb, 1990), transformation (Simoes and Costa, 2001), adaptive crossover and mutation probabilities (Wong et al., 2003), adaptive selection probability (Shimodaira, 2001), random immigrant (Grefenstette, 1992), and so forth.

Other methods involve the balance between exploration and exploitation by explicit controlling the population diversity. Ursem proposed the diversity-guided evolutionary algorithm (DGEA), in which a distance-to-average-point diversity measure is controlled to alternate between phases of exploration and exploitation (Ursem, 2002). Riget and Vesterstrøm used the distance-to-average-point diversity measure to control the search of the PSO algorithm (Riget and Vesterstrøm, 2002). Lu and Traoré proposed an evolutionary algorithm with a new entropy-based fitness function to estimate the optimal number of data clusters in cluster analysis (Lu and Traore, 2005). In (Liu et al., 2007), Liu et al. presented an entropy-driven parameter control method to balance the exploration and the exploitation in evolutionary algorithms. Nguyen and Wong used control theory to adjust the mutation rate in a unimodal space according to the desired diversity, which varies exponentially with time steps (Nguyen and Wong, 2003). In their method, when the current population diversity deviates from the desired one, the mutation rate is adjusted as a control problem. In (Gouvea Jr. and Araugo, 2007), the authors proposed an evolutionary algorithm with diversity-reference adaptive control based on a model-reference adaptive system.

## 3 The QPSO Algorithm

In the original PSO with $M$ individuals, each individual is treated as a volume-less particle in the $N$-dimensional space, with the position vector and velocity vector of particle $i$ at $n^{\text{th}}$ iteration represented as $X_{i,n} = (X_{i,n}^1, X_{i,n}^2, \cdots, X_{i,n}^N)$ and $V_{i,n} = (V_{i,n}^1, V_{i,n}^2, \cdots, V_{i,n}^N)$. Vector $P_{i,n} = (P_{i,n}^1, P_{i,n}^2, \cdots, P_{i,n}^N)$ is the best previous position (the position giving the best objective function value since the initialization of the population) of particle $i$ called the *personal best* (*pbest*) position, and vector $G_n = (G_n^1, G_n^2, \cdots, G_n^N)$ is the position of the best particle among all the particles in the population found so far and called the *global best* (*gbest*) position. Without loss of generality, we consider the following minimization problems

$$\text{Minimize} \quad f(X), \quad s.t. \quad X \in S \subseteq R^N, \tag{1}$$

where $f(X)$ is an objective function and $S$ is the feasible space. Accordingly, $P_{i,n}$ can be updated by

$$P_{i,n} = \begin{cases} X_{i,n} & \text{if} \quad f(X_{i,n}) < f(P_{i,n-1}) \\ P_{i,n-1} & \text{if} \quad f(X_{i,n}) \geq f(P_{i,n-1}) \end{cases}. \tag{2}$$

Since $G_n$ is the *pbest* position giving the best fitness value among all particles, it can be found by

$$G_n = P_{g,n}, \quad g = \arg\min_{1 \leq i \leq M}[f(P_{i,n})]. \tag{3}$$

With the above definitions, each particle in the original PSO updates its velocity and position by the following equations:

$$V_{i,n+1}^j = V_{i,n}^j + c_1 r_{i,n}^j (P_{i,n}^j - X_{i,n}^j) + c_2 R_{i,n}^j (G_n^j - X_{i,n}^j), \tag{4}$$

$$X_{i,n+1}^j = X_{i,n}^j + V_{i,n+1}^j, \tag{5}$$

for $i = 1, 2, \cdots M; j = 1, 2 \cdots, N$, where $c_1$ and $c_2$ are known as the acceleration coefficients. The parameters $r_{i,n}^j$ and $R_{i,n}^j$ are two different random numbers distributed uniformly within (0, 1), that is $r_{i,n}^j, R_{i,n}^j \sim U(0,1)$. Generally, the value of $V_{i,n}^j$ is restricted on the interval $[-V_{\max}, V_{\max}]$.

Clerc and Kennedy undertook a formal analysis of the particle's trajectory and the stability properties of the PSO algorithm (Clerc and Kennedy, 2002). They simplified the particle swarm to a second-order linear dynamical system, whose stability depends on the system poles or the eigenvalues of the state matrix. Their analyses essentially revealed the fact that the particle swarm may converge if each particle converges toward its local focus, $p_{i,n} = (p_{i,n}^1, p_{i,n}^2, \cdots p_{i,n}^N)$ defined at the coordinates

$$p_{i,n}^j = \frac{c_1 r_{i,n}^j P_{i,n}^j + c_2 R_{i,n}^j G_n^j}{c_1 r_{i,n}^j + c_2 R_{i,n}^j}, \quad 1 \leq j \leq N, \tag{6}$$

or

$$p_{i,n}^j = \varphi_{i,n}^j P_{i,n}^j + (1 - \varphi_{i,n}^j) G_n^j, \tag{7}$$

where $\varphi_{i,n}^j = c_1 r_{i,n}^j / (c_1 r_{i,n}^j + c_2 R_{i,n}^j)$ with regard to the random numbers $r_{i,n}^j$ and $R_{i,n}^j$ defined in equations (4), (6). The acceleration coefficients $c_1$ and $c_2$ in the original PSO are generally set to be equal, i.e. $c_1 = c_2$, and if so, $\varphi_{i,n}^j$ will be a sequence of uniformly distributed random numbers within (0,1). As a result, equation (7) can be restated as

$$p_{i,n}^j = \varphi_{i,n}^j P_{i,n}^j + (1 - \varphi_{i,n}^j) G_n^j, \quad \varphi_{i,n}^j \sim U(0,1). \tag{8}$$

The above equation indicates that $p_{i,n}$ is a stochastic point that lies in a hyper-rectangle with $P_{i,n}$ and $G_n$ being two ends of its diagonal, and that moves following the $P_{i,n}$ and $G_n$. In the process of convergence, the particle moves around and careens toward point $p_i$ with its kinetic energy (velocity) declining to zero, like a returning satellite orbiting the earth. As such, a particle in PSO can be considered as flying in an attraction potential field centered at point $p_{i,n}$ in Newtonian space. It has to be in bound state for the sake of avoiding explosion and guaranteeing convergence. If these conditions are generalized to the case when the particle in PSO moves in quantum space, it is also indispensable that the particle moves in a quantum potential field to ensure the bound state. From the perspective of quantum mechanics, the bound state in quantum space, however, is entirely different from that in Newtonian space, which may lead to a very different form of PSO. This is the motivation of the proposed QPSO algorithm (Sun et al., 2004a).

In QPSO, each single particle is assumed to be a spin-less one, with quantum behavior. Thus, the state of the particle is characterized by a wavefunction $\psi$, where $|\psi|^2$ is the probability density function of its position. At the *n*th iteration, particle *i* moves in the *N*-dimensional space with a $\delta$ potential well centered at $p_{i,n}^j$ on the *j*th dimension for $1 \le j \le N$. If $Y_{i,n+1}^j = |X_{i,n+1}^j - p_{i,n}^j|$, we can get the following normalized wavefunction at the (*n*+1)th iteration satisfying the bound condition that $\psi(Y_{i,n+1}^j) \to 0$ as $Y_{i,n+1}^j \to \infty$.

$$\psi(Y_{i,n+1}^j) = \frac{1}{\sqrt{L_{i,n}^j}} \exp(-Y_{i,n+1}^j / L_{i,n}^j), \tag{9}$$

where $L_{i,n}^j$ is the characteristic length of the wave function. By the definition of the wavefunction, the probability density function is given by:

$$Q(Y_{i,n+1}^j) = |\psi(Y_{i,n+1}^j)|^2 = \frac{1}{L_{i,n}^j} \exp(-2Y_{i,n+1}^j / L_{i,n}^j), \tag{10}$$

and, thus, the probability distribution function is

$$F(Y_{i,n+1}^j) = 1 - \exp(-2Y_{i,n+1}^j / L_{i,n}^j). \tag{11}$$

Using the Monte Carlo method, we can obtain the *j*th component of the position of particle *i* at the (*n*+1)th iteration by

$$X_{i,n+1}^j = p_{i,n}^j \pm \frac{L_{i,n}^j}{2} \ln(1/u_{i,n+1}^j), \quad u_{i,n+1}^j \sim U(0,1), \tag{12}$$

where $u_{i,n}^j$ is a sequence of random numbers uniformly distributed in (0, 1). As proposed in (Sun et al, 2004a, 2004b), the value of $L_{i,n}^j$ is given either by

$$L_{i,n}^j = 2\alpha |X_{i,n}^j - p_{i,n}^j|, \tag{13}$$

or

$$L_{i,n}^j = 2\alpha |X_{i,n}^j - C_n^j|, \tag{14}$$

where $C_n = (C_n^1, C_n^2, \cdots, C_n^N)$ is known as the mean best (*mbest*) position, defined by the average of the *pbest* positions of all particles, that is, $C_n^j = (1/M) \sum_{i=1}^{M} P_{i,n}^j \quad (1 \le j \le N)$. Therefore, the position of the particle can be updated according to

$$X_{i,n+1}^j = p_{i,n}^j \pm \alpha |X_{i,n}^j - p_n^j| \ln(1/u_{i,n+1}^j), \tag{15}$$

or

$$X_{i,n+1}^j = p_{i,n}^j \pm \alpha |X_{i,n}^j - C_n^j| \ln(1/u_{i,n+1}^j). \tag{16}$$

The parameter $\alpha$ in (13) and (14) is called the contraction-expansion (CE) coefficient, which can be adjusted to control the convergence speed of the particle. The equations (15) and (16) lead to two versions of

J. Sun, X. Wu, V. Palade, W. Fang, Z. Wang

QPSO, which are denoted as QPSO-Type 1 and QPSO-Type 2. Since the QPSO-Type 2 generally has a better performance than the QPSO-Type 1, it has been widely used and accepted as the standard QPSO algorithm. In this paper, our theoretical and empirical analyses aim at this QPSO version. The procedure of the standard QPSO is outlined below in Algorithm 1, where $\text{rand}i(\cdot)$, $i=1,2,3$ are separately generated random numbers uniformly distributed in (0,1).

---

**Algorithm 1: The QPSO Algorithm**

**Begin**
  Initialize the current position $X_{i,0}^j$ and the personal best position $P_{i,0}^j$ of each particle, evaluate their fitness values and find the global best position $G_0$; Set $n=0$.
  **While** (termination condition = false)
  **Do**
    Set $n=n+1$;
    Compute mean best position $C_n$ and select the value of $\alpha$ properly;
    **for** ($i=1$ to $M$)
      **for** ($j=1$ to $N$)
        $\varphi_{i,n}^j = \text{rand}1(\cdot)$;
        $p_{i,n}^j = \varphi_{i,n}^j P_{i,n}^j + (1-\varphi_{i,n}^j)G_n^j$;
        $u_{i,n}^j = \text{rand}2(\cdot)$;
        **if** ( $\text{rand}3(\cdot) < 0.5$ )
          $X_{i,n+1}^j = p_{i,n}^j + \alpha |X_{i,n}^j - C_n^j| \ln(1/u_{i,n+1}^j)$;
        **else**
          $X_{i,n+1}^j = p_{i,n}^j - \alpha |X_{i,n}^j - C_n^j| \ln(1/u_{i,n+1}^j)$;
        **end if**
      **end for**
      Evaluate the fitness value of $X_{i,n+1}$, i.e. the objective function value $f(X_{i,n+1})$;
      Update $P_{i,n}$ and $G_n$;
    **end for**
  **end do**
**end**

---

## 4. Diversity Analysis for the QPSO

### 4.1 Measuring Diversity

For a real-value encoded evolutionary algorithm, the structural difference is represented by the distance between the two vectors. The genotype diversity, in this case, can be measured by the distance to the average point as suggested in (Ursem, 2002). Since the phenotype diversity counts the number of unique fitness values in a population, those used in other evolutionary algorithms, such as binary GAs and GP, are not applicable to the real-value encoded algorithm. Here, we adopt proportional entropy of the fitness values as the phenotype diversity. This is defined by

$$S = -\sum_i q_i \log_2 q_i , \qquad (17)$$

where $q_i$ is the proportion of the population occupied by the population partition $i$ (Liu et al, 2007). A partition is assumed to be each possible different fitness value, but could be defined to include a subset of values. In the proportional entropy, $q_i$ is formalized as $f_i \Big/ \sum_{i=1}^{M} f_i$, where $f_i$ is the fitness value of an individual, $M$ is the population size and $q_i$ is the criterion for categorizing fitness classes. In the continuous real-space, since the probability that two individuals have the same fitness value is practically zero, we can assume that all individuals of a population have different fitness values, namely, different $q_i$ values. The number of fitness classes then equals the population size.

We use the above two measures for the population diversity of in the QPSO algorithm. However, unlike other real-value encoded evolutionary algorithms, there are two sets of position vectors in the QPSO at each iteration - the set of particles' current position vectors, $X_n = (X_{1,n}, X_{2,n}, \cdots, X_{M,n})$; and the one of particles' personal best position vectors, $P_n = (P_{1,n}, P_{2,n}, \cdots, P_{M,n})$. Correspondingly, we have two distance-to-average-point diversities for $X_n$ and $P_n$, respectively, which are defined by

$$D(X_n) = \frac{1}{M \cdot A} \sum_{i=1}^{M} [\sum_{j=1}^{N} [X_{i,n}^j - \bar{X}_n^j]^2]^{1/2} = \frac{1}{M \cdot A} \sum_{i=1}^{M} |X_{i,n} - \bar{X}_n|, \qquad (18)$$

and

$$D(P_n) = \frac{1}{M \cdot A} \sum_{i=1}^{M} [\sum_{j=1}^{N} [P_{i,n}^j - C_n^j]^2]^{1/2} = \frac{1}{M \cdot A} \sum_{i=1}^{M} |P_{i,n} - C_n|, \qquad (19)$$

where $A$ is the length of the longest diagonal in the search space, $\bar{X}_n^j$ and $\bar{P}_n^j$ are the $j$'th values of the average points of $X_n$ and $P_n$, that is, $\bar{X}_n^j = (1/M) \sum_{i=1}^{M} X_{i,n}^j$ and $\bar{P}_n^j = (1/M) \sum_{i=1}^{M} P_{i,n}^j$. There are also two entropy diversities:

$$S(X_n) = -\sum_{i=1}^{M} q(X_{i,n}) \log_2 q(X_{i,n}), \qquad (20)$$

and

$$S(P_n) = -\sum_{i=1}^{M} q(P_{i,n}) \log_2 q(P_{i,n}), \qquad (21)$$

where $q(X_{i,n}) = f(X_{i,n}) / \sum_{i=1}^{M} f(X_{i,n})$ and $q(P_{i,n}) = f(P_{i,n}) / \sum_{i=1}^{M} f(P_{i,n})$.

**4.2 Correlation Measure between Fitness and Diversity**

One of the objectives of this paper is to quantify the importance and levels of diversity, recorded by the two measures, on several typical benchmark functions. The relationship between diversity and fitness is measured by the Spearman correlation, which ranks two sets of variables and tests for a linear relationship between the ranks of variables. The Spearman correlation coefficient is computed as follows:

$$1 - \frac{6 \sum_{j=1}^{J} d_i^2}{J^3 - J}, \qquad (22)$$

where $J$ is the number of items, and $d_i$ is the distance between each population's rank of fitness and rank of diversity. A value of -1.0 represents a negative correlation, 0.0 denotes no correlation, and 1.0 indicates positive relation. For our measures, a negative correlation implies that either good fitness accompanies high diversity or bad fitness accompanies low diversity. Alternatively, a positive correlation indicates that either good fitness accompanies low diversity or bad fitness accompanies high diversity. All of the testing results are presented in the following subsection.

**4.3 Results and Analysis**

The experiments on diversity measures were performed on several widely used benchmark functions, including Sphere, Rosenbrock, Rastrigin and Griewank, whose mathematical expression are listed in Table 1. The dimension of each problem is five and the algorithm was executed for 500 iterations in a single run on each problem. In (Sun et al., 2012b), it is suggested that if time-varying $\alpha$ is used, the QPSO generates good performance by decreasing its value linearly from 1.0 to 0.5, while if $\alpha$ is set at a fixed value, then $\alpha = 0.75$ yields good results in general. Thus, in our experiments, we tested the algorithm by using 20 particles and both of the controlling methods for $\alpha$ with the above recommended parameter settings.

Table 1: The four benchmark functions for testing diversity measures

| Sphere Function | $f_1(X) = \sum_{i=1}^{N} x_i^2$ |
|---|---|
| Rosenbrock Function | $f_2(X) = \sum_{i=1}^{N-1} (100 \cdot (x_{i+1} - x_i^2)^2 + (x_i - 1)^2)$ |
| Rastrigin Function | $f_3(X) = \sum_{i=1}^{N} (x_i^2 - 10 \cdot \cos(2\pi x_i) - 10)$ |
| Griewank Function | $f_4(X) = (1/4000) \sum_{i=1}^{N} x_i^2 - \prod_{i=1}^{N} \cos(x_i / \sqrt{i}) + 1$ |

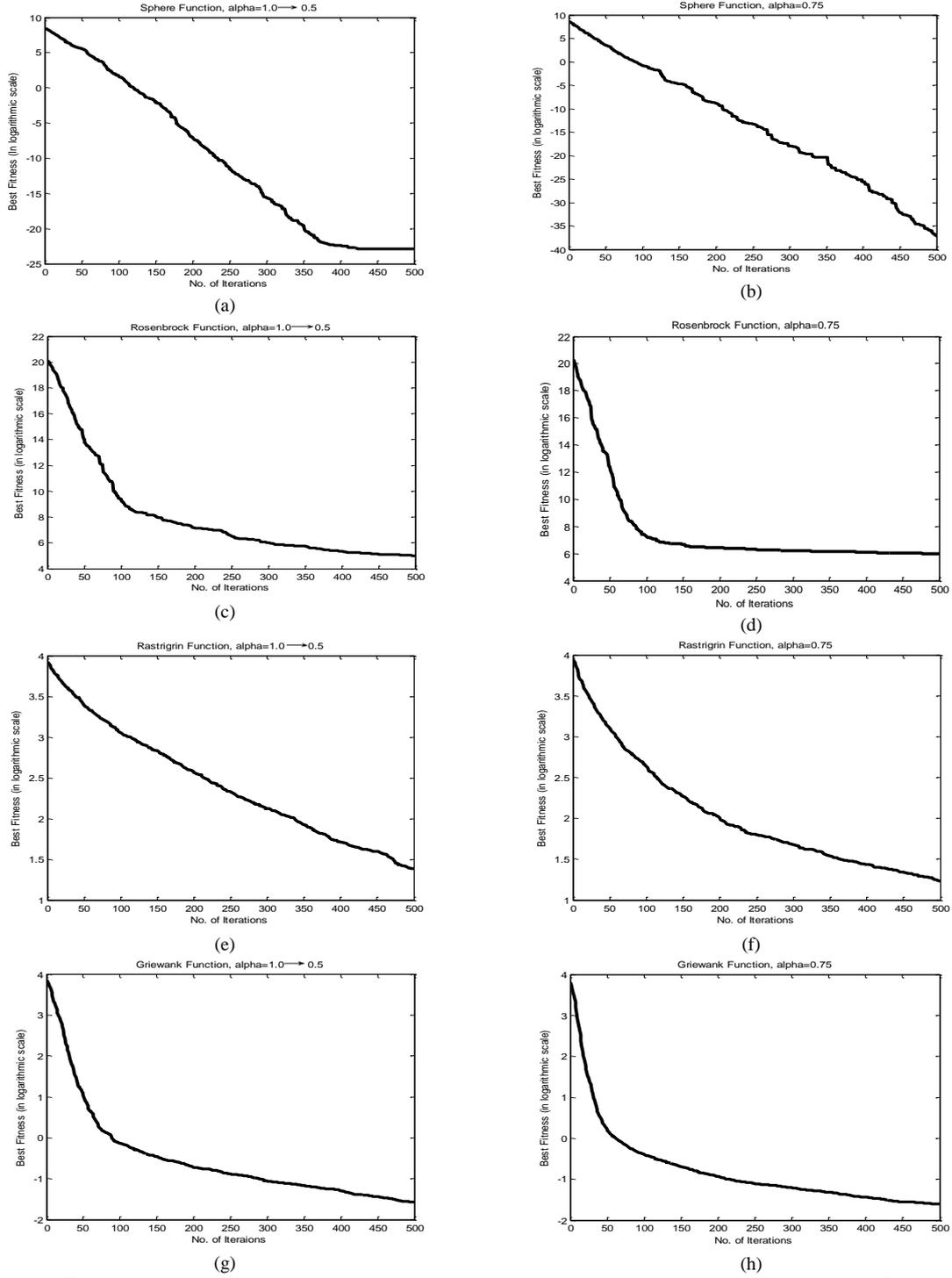

Figure 1. The best fitness value (in logarithmic scale) at each iteration for the Sphere, Rosenbrock, Rastrigin and Griewank functions with dimension=5, plotted against the iteration number. (Average value of the best fitness over 100 independent runs for each problem)

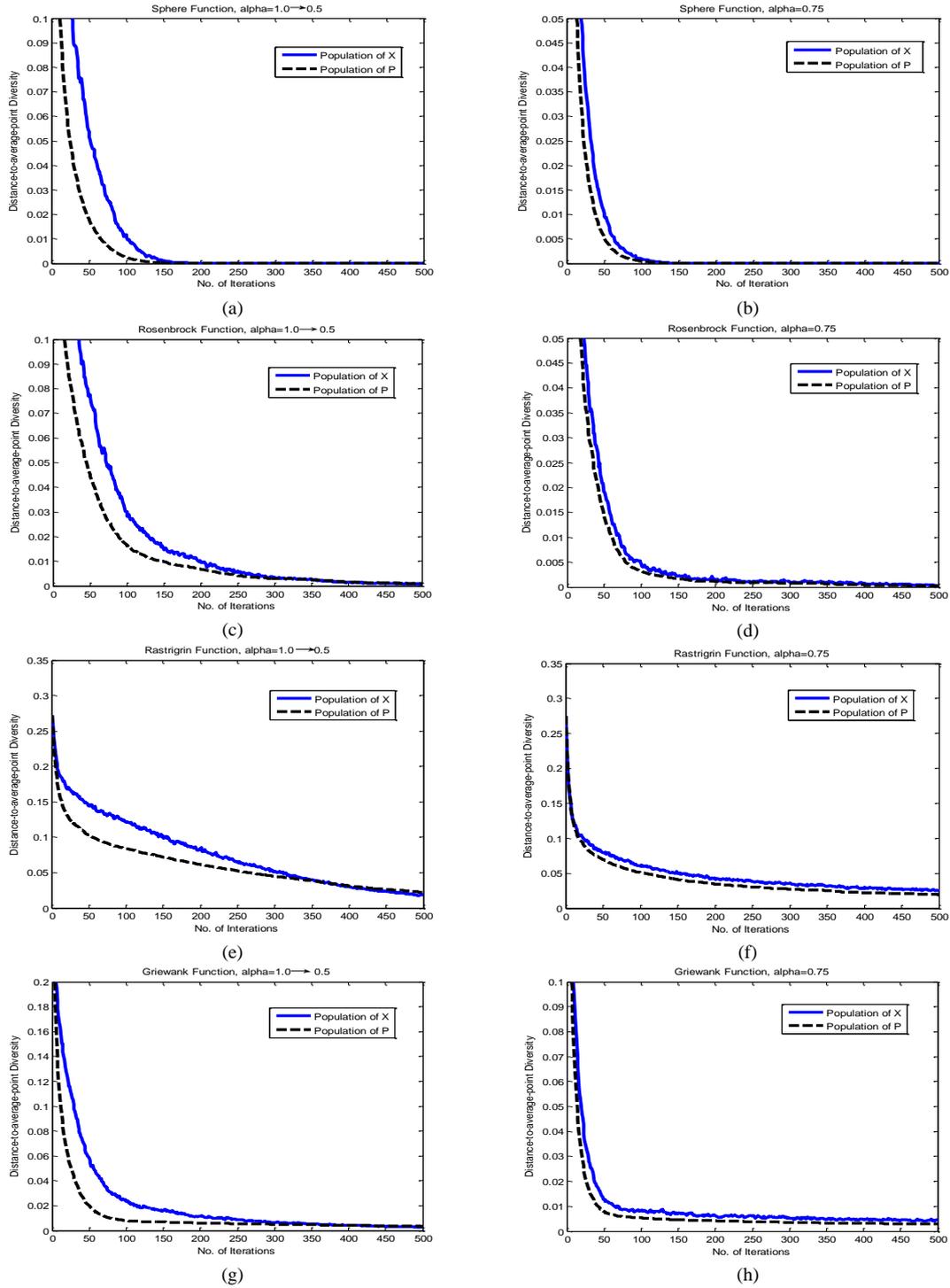

Figure 2. The distance-to-average-point diversities at each iteration for the Sphere, Rosenbrock, Rastrigin and Griewank functions, plotted against the iteration number. (Average value of the diversity over 100 independent random runs for each problem)

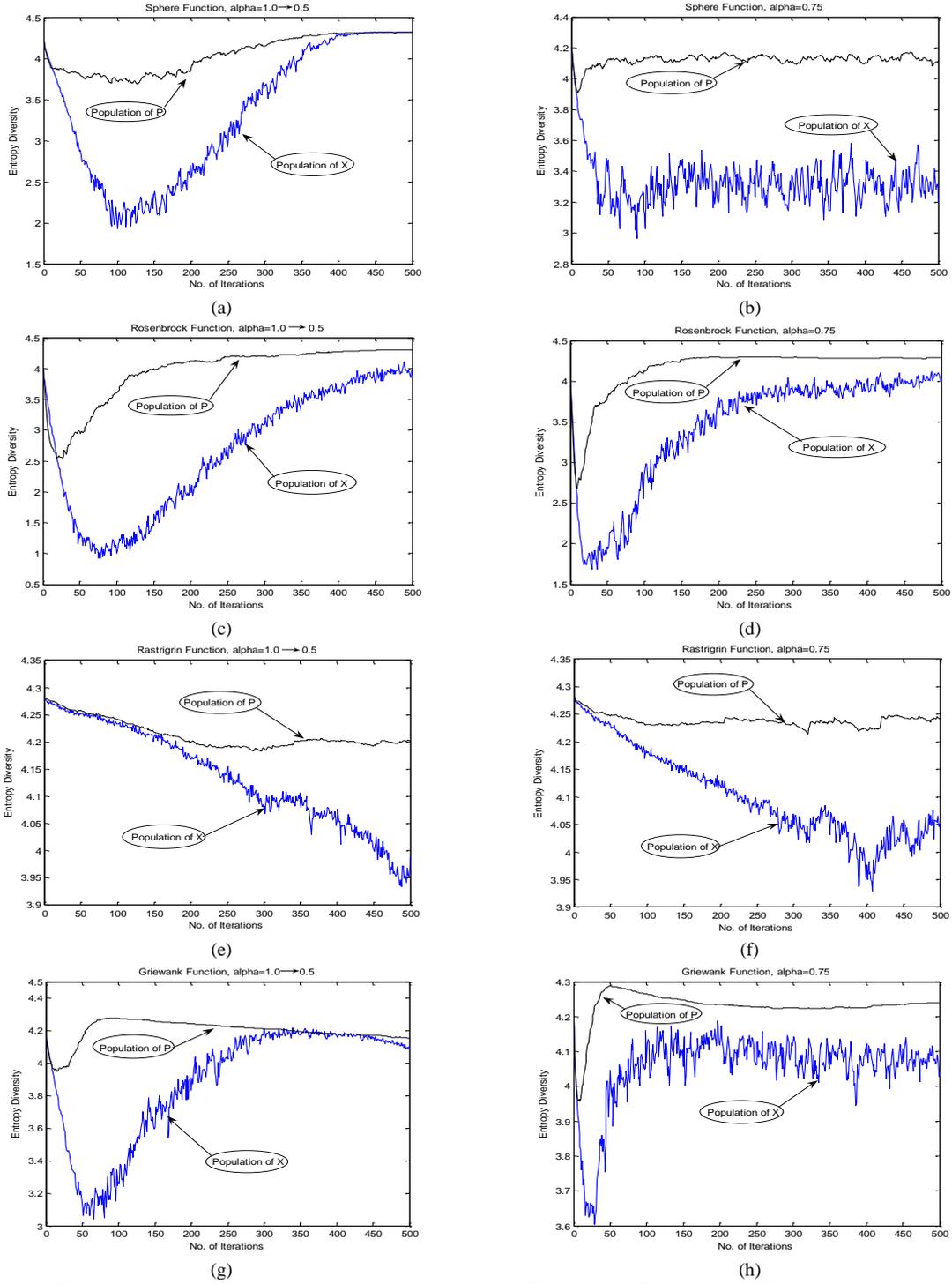

Figure 3. The entropy diversities at each iteration for the Sphere, Rosenbrock, Rastrigin and Griewank functions, plotted against the iteration number. (Average value of the diversity over 100 independent random runs for each problem).

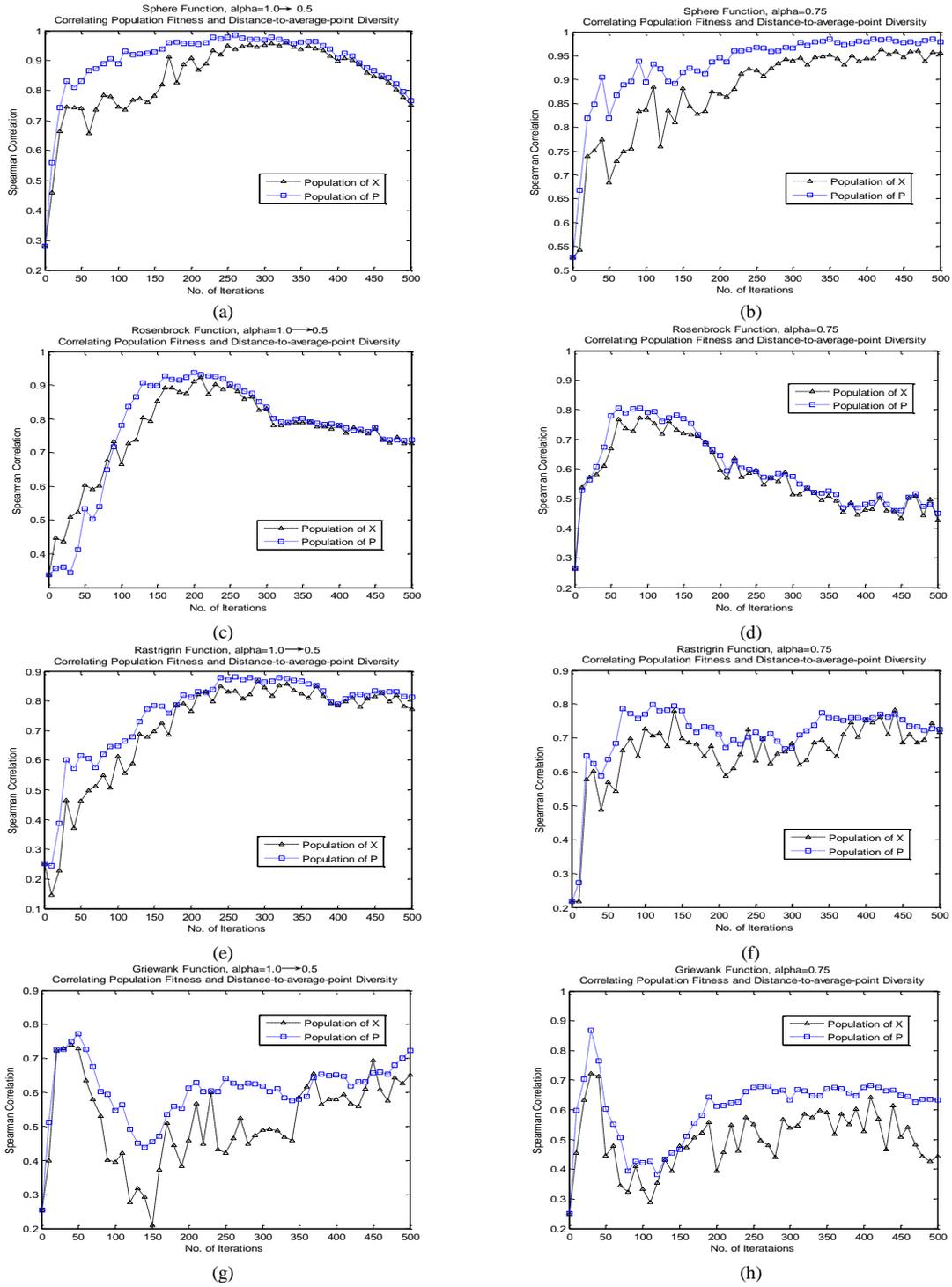

Figure 4. Correlations between the best fitness and the distance-to-average-point diversities. Each point represents the correlation among 100 populations selected from 100 independent runs at every 10 iterations)

J. Sun, X. Wu, V. Palade, W. Fang, Z. Wang

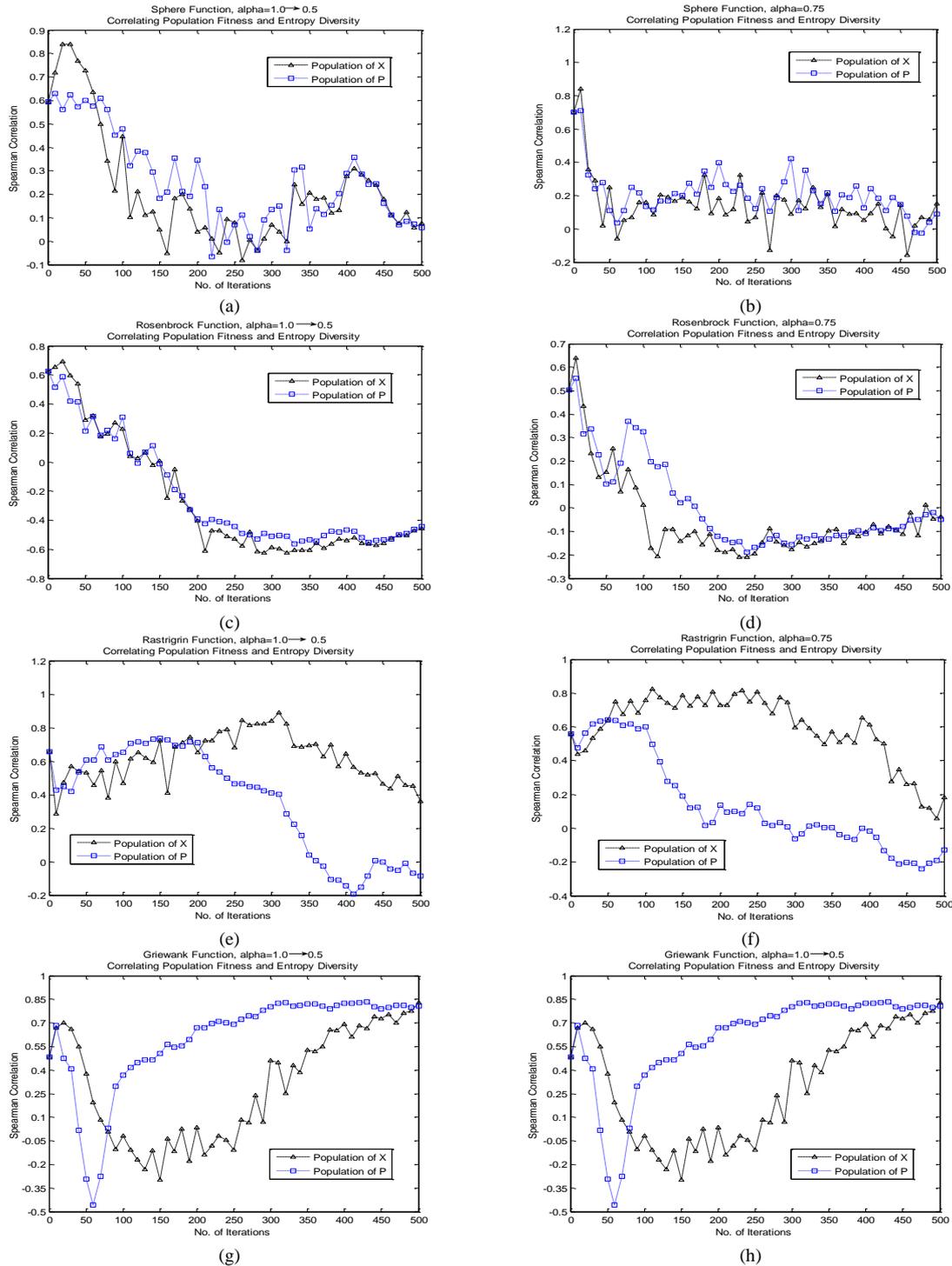

Figure 5. Correlations between the best fitness and the entropy diversities. Each point represents the correlation among 100 populations selected from 100 independent runs at every 10 iterations)

Firstly, we examine the experimental results, by focusing on the trends that populations exhibit on the best fitness (the fitness of the global best position) and the diversity measures. We then attempt to present a more general analysis of how effective the two diversity measures are, and what diversity tells us about evolving populations, by investigating the correlations between the best fitness and the diversity measures in evolving swarms.

We investigated 100 independent runs for each problem, for the selected diversity measures and for each parameter control method. Figure 1 shows the best fitness (averaged over 100 runs and in logarithmic scale) at each iteration during the search process. Figures 2 and 3 trace the distance-to-the-average-point and entropy diversities at each iteration of the search process in each case, respectively. The results visualized in the two figures are averaged over the 100 runs. Figure 4 shows the correlation between the distance-to-the-average-point diversities and best fitness for each iteration, while Figure 5 presents the correlation between the entropy diversities and best fitness. From Figure 2, it is clear that the distance-to-the-average-point diversities of $X_n$ and $P_n$ decline to zero during the course of the search, which supports the results of Theorem 5. On the other hand, as shown in Figure 2, the entropy diversities reduced at the early stage, reached the bottom, and then bounced to a higher level and approached $\log_2(M) = \log_2 20 \approx 4.3219$ in most cases.

For the Sphere function, Figure 1(a) shows that when the time-varying $\alpha$ was used, the improvement of best fitness value seems to stagnate after 400 iterations. Correspondingly, the distance-to-the-average-point diversities of $X_n$ and $P_n$ reached zero rapidly, while their entropy diversities decreased at the early stage of the search process and, at the later stage, both increased to $\log_2 20$, as shown by Figures 2(a) and 3(a), respectively. Figure 4(a) presents the correlations between the distance-to-average-point diversity and the best fitness of $X_n$ and $P_n$ across 500 iterations over 100 runs. It can be observed that each of the correlation coefficients started at a relatively small positive value at initialization and then increased to a value close to +1. After 400 iterations, the correlation coefficient decreased, which was accompanied by the stagnation of the improvement of the best fitness value. The correlations between the entropy diversities and the best fitness decreased from positive values to negative ones and, then, experienced an oscillative ascendance. When $\alpha$ was fixed at 0.75 during the search, the best fitness value constantly improved without stagnation within 500 iterations, as shown in Figure 1(b), and the distance-to-average-point diversities converged towards zero but not so fast as in the case of time-varying $\alpha$, as evident from Figure 2(b). Figure 3(b) indicates that the entropy diversity of $P_n$ was maintained at a value around 4.1, while the entropy diversity of $X_n$ decreased from 4.2 to around 3.0, and then was in fluctuation until the end of the search. The correlation coefficients between the distance-to-average-point diversities and the best fitness, which can be visualized in Figure 4(b), increased from around 0.53 to 1 without remarkable decline. The correlations showed in Figure 5(b) between the entropy diversities and the best fitness value, however, varied in a similar pattern with those when time-varying $\alpha$ was used.

In Figure 1(c), the improvement of the best fitness for the Rosenbrock function decreased and stagnated after 150 iterations, when time-varying $\alpha$ was used. The correlations between the distance-to-average-point diversities and the best fitness reached their highest levels between the 150th and 200th iteration, and decreased since then, accompanying the decrease and stagnation of the improvement of the best fitness value, as shown in Figure 4(c). At the same time, Figure 5(c) shows that the correlations between the entropy diversities and the best fitness decreased from about 0.6 to -0.6, and increased slightly after 350 iterations. If the parameter $\alpha$ was fixed during the search, Figure 4(d) shows that the improvement of the best fitness value experienced a gradual stagnation, which was accompanied by the reduction of the correlations between the distance-to-average diversities and the best fitness value after about 100 iterations of oscillative increase.

As for the Rastrigin function, it is shown in Figure 1(e) that with the time-varying $\alpha$, the best fitness value experience continuous improvement without stagnation or significant slowdown, which is accompanied by a sustained rise of the correlations between the distance-to-average-point diversities and the best fitness value, as shown in Figure 4(e). For the QPSO with the fixed $\alpha$, there was a slight slowdown of the improvement of the best fitness and, correspondingly, the correlations between the distance-to-average diversities and the best fitness had a slow decrease in oscillation, as shown in Figure 4(f). On the other hand, the entropy diversities and their correlations with the best fitness value showed little association with the improvement process of the best fitness.

The results for the Griewank function in Figure 1(g) and Figure 1(h) show that there was some slowdown in the improvement of the best fitness value after 50 iterations and 40 iterations for the QPSO with the time-varying $\alpha$ and fixed $\alpha$, respectively. The correlations between the distance-to-average-point diversities and the best fitness also showed obvious decreases, as shown in Figures 4(g) and 4(h). As for the entropies and their correlations with the best fitness, they also showed little association with the change of the improvement of the best fitness value.

From the above analysis, we find that the distance-to-average-point diversities declined constantly and their correlations with the best fitness were positive throughout all of the evolving processes, while the entropy diversities showed to be a bit fickle, and their correlations with the best fitness value were alternately positive or negative. Juxtaposing the improvement process of the best fitness with the dynamical changes of the correlations between the distance-to-average-point diversities and the best fitness value, we find that there is a strong association between the change in the improvement of the best fitness and the trends of the correlation coefficients. On the other hand, there is little evidence for the association of the variations of the best fitness improvement with the trends of the correlations between entropy diversities and the best fitness. Therefore, it can be concluded that the distance-to-average-point diversities may play a more important role in evolving populations than the entropy diversities. It may be because there are no operators based on the fitness values in QPSO, such as the selection operators in EAs.

The positivity of the correlation coefficients between the distance-to-average-point diversities and the best fitness reflects the fact that good fitness accompanies low diversities of $X_n$ and $P_n$. Therefore, low diversities are not necessarily the reason for poor performance. In fact, from the above analysis, we have found that at the early stage of the evolving process, the best fitness improved rapidly, as the correlation coefficients increased, and the diversities decreased rapidly as well. This means that rapid decline of the distance-to-average-point diversities is desirable for the improvement of the fitness. During the search process, if the fitness improvement encounters a slowdown or even stagnation, the correlation coefficients could experience a decline, which implying that the diversities may decrease too fast so that the improvement of the best fitness slows down. The faster the correlations decrease, the greater the slowdown of the improvement of the best fitness. As such, it is obvious that viewing the trends of the correlation between diversities and the best fitness, instead of only focusing on the values of the correlations, is of vital importance for grasping the properties of the evolving process. The ideal situation is when the correlation coefficients increase constantly and approach 1, as in the case of the Sphere function with $\alpha = 0.75$. In reality, this ideal situation can not be acquired artificially. A practical measurement is to find some efficient strategies to control the distance-to-average-point diversities, as we will do in the next section.

## 5   Controlling the Diversity

The diversity control strategies designed in this work focus on the distance-to-average-point diversities, since they have been identified to play a more important role than the entropy diversities, and their variation is a vital factor of the algorithmic performance during the search process of the QPSO. In this section, two diversity control strategies are proposed for improving the search performance of the QPSO.

### 5.1 The Three-Phased Diversity Control Strategy

It has been found in the previous section that, in the QPSO, good fitness value of the *gbest* position is associated with low distance-to-average-point diversities, and the rapid decline of the diversities at the early stage of the search process is desirable for the improvement of the fitness values. During the evolving process, the improvement of the fitness can decrease or even stagnate accompanied by the reduction of the correlation between the diversities and the best fitness. This reveals the fact that excessively low levels of the diversities resulted from their rapid decreases account for the stagnation or premature convergence of the QPSO algorithm. As such, maintaining the diversities at a certain level can be an effective way to avoid premature convergence of the QPSO algorithm.

The proposed strategy controls explicitly the diversity of $X_n$ by setting an upper bound $d_{upper}$ and a lower bound $d_{lower}$ for the diversity and it divides the search mode of the QPSO algorithm into three phases. Phase 1 starts at the initialization of the swarm and ends when the diversity first reaches $d_{lower}$. The search of the QPSO algorithm in phase 1 is essentially identical to the search process of the standard QPSO with $\alpha = \alpha_1 < e^{\gamma}$. In this phase, the diversities of $X_n$ and $P_n$ decline and their decreasing speeds are mutually influenced. That is, the QPSO in phase 1 runs in a convergence mode. As indicated in Section 4, the convergence of each $P_{i,n}$ to the *gbest* position $G_n$ is determined by not only the distribution of $X_{i,n}$, but also the level set of $P_{i,n}$ as defined in Section 4, which is mainly dependent on the landscape of the objective function. For some problems, $P_{i,n}$ may converge very slowly even though the value of the CE coefficient $\alpha$ is very small. A slow convergence of $P_{i,n}$ in turn leads to a slow convergence of $X_{i,n}$. From the viewpoint of diversities, the diversities of $P_n$ for these problems may have slow decreasing speed so that the phase 1 could not end when the whole search process is over. Too slow decline of the diversities may not be conducive to the search performance of the algorithm as has been addressed. Thus, in our proposed strategy, an upper bound $n_{phase\_1}$ is set for the number of iterations for which the algorithm runs in phase 1. If the algorithm has run for $n_{phase\_1}$ iterations and it is still in phase 1, set the *pbest* position $P_{i,n}$ to be its local focus $p_{i,n}$ so that $P_{i,n}$ moves to $G_n$ compulsorily and the diversity of $P_n$ declines rapidly. As a result, the diversity of $X_n$ also decreases rapidly until it reaches $d_{lower}$.

After phase 1, the QPSO runs alternatively in phase 2 and phase 3. In phase 2, the particle swarm explodes in order to increase the diversity of $X_n$ by setting the CE coefficient $\alpha = \alpha_2 > e^\gamma$ until it reaches the upper bound $d_{upper}$. Then comes phase 3, in which the CE coefficient is set to be $\alpha = \alpha_3 < e^\gamma$ so that the QPSO in this phase searches in a convergence mode. Alternation of phase 2 and phase 3 after phase 1 is similar to the two-phased strategy proposed for the PSO in (Riget and Vesterstrøm, 2002), which was inspired by the work in (Ursem, 2002). However, the two-phased PSO in (Riget and Vesterstrøm, 2002) did not take into account the interplay between the diversities of $X_n$ and $P_n$, which may affect the effectiveness of the diversity controlling strategy.

The procedure of the QPSO with the three-phased diversity control strategy (QPSO-TDC) is outlined below in Algorithm 2.

---

**Algorithm 2: The QPSO-TDC Algorithm**

**Begin**
Initialize the current position $X_{i,0}^j$ and the personal best position $P_{i,0}^j$ of each particle, evaluate their fitness values and find the global best position $G_0 = P_{g,0}$;
Pre-assign $\alpha_2 > 1.781$ and $\alpha_3 < 1.781$ and the parameter setting for $\alpha_1$. Pre-set $d_{lower}$, $d_{upper}$ and $n_{phase\_1}$;
Phase=1 (in Phase 1);
**for** $n = 1$ to $n_{\max}$
    Compute the mean best position and calculate $D(X_n)$ by using (49);
  **if** ($D(X_n) < d_{lower}$ and Phase=1)
    Phase=2; (in Phase 2)
  **endif**
  **if** ($D(X_n) > d_{upper}$ and Phase=2)
    Phase=3; (in Phase 3)
  **endif**
  Select the value of $\alpha$ to be $\alpha_1$, $\alpha_2$ and $\alpha_3$ corresponding to Phase 1, Phase 2 and Phase 3, respectively;
  **for** ($i$=1 to $M$)
    **for** $j$=1 to $N$
      $\varphi_{i,n}^j = \text{rand1}(\cdot)$;
      $p_{i,n}^j = \varphi_{i,n}^j P_{i,n}^j + (1-\varphi_{i,n}^j) G_n^j$;
      $u_{i,n}^j = \text{rand2}(\cdot)$;
      **if** ($\text{rand3}(\cdot) < 0.5$)
        $X_{i,n+1}^j = p_{i,n}^j + \alpha \mid X_{i,n}^j - C_n^j \mid \ln(1/u_{i,n+1}^j)$;
      **else**
        $X_{i,n+1}^j = p_{i,n}^j - \alpha \mid X_{i,n}^j - C_n^j \mid \ln(1/u_{i,n+1}^j)$;
      **end if**
    **end for**
    Evaluate the objective function value $f(X_{i,n})$;
    Update $P_{i,n}$;
    **If** (Phase=1 and $n > n_{phase\_1}$)
      $P_{i,n} = p_{i,n}$;
    **end if**
    Update $G_n$;
  **end for**
**end for**
**end**

---

### 5.2 Controlling the Declining Speed of the Diversity

The second proposed strategy here aims at controlling the declining speed of the distance-to-average-point diversity. As it has been observed in Section 5, decreasing the diversities at the early stage of the evolving process is beneficial for the improvement of the fitness value, which is reflected by the increase of the correlations between the diversities and the best fitness. However, excessively rapid decrease of the diversities can result in the decrease or even the stagnation of the improvement of fitness. On the other hand, for some problems, undesirably slow decline of the diversities may also lead to poor search performance as indicated in the previous subsection. This brings us the motivation of controlling the declining speed of the diversity throughout the whole evolving process.

For some unimodal functions, it is desirable to decline the diversity in an exponential way for a good algorithmic performance, which has been verified by the results for the Sphere function when $\alpha = 0.75$, for the QPSO. Nguyen and Wong also defined the desired diversity of EAs in a unimodal space as the one that exponentially decreases with generations (Nguyen and Wong, 2003). For most of the multimodal problems, the exponential decline of the diversities may be so fast that the decrease and stagnation of the fitness improvement are prone to be encountered, as show in Section 5. In the proposed approach, the desired

J. Sun, X. Wu, V. Palade, W. Fang, Z. Wang

diversity of $X_n$ at the *n*th iteration is defined as the one which decreases no faster than polynomially in terms of the iteration number; that is:

$$D_{d,n} = \frac{(n_{\max} - n)^r}{(n_{\max})^r}(D_{d,initial} - D_{d,final}) + D_{d,final}, \tag{57}$$

where $D_{d,n}$ is the desired diversity at the *n*th iteration, which essentially gives the lower bound for the diversity at the *n*th iteration, and $n_{\max}$ is the maximum number of iterations. Parameter *r* is a positive number and is generally set to be larger than 1, $D_{d,initial}$ and $D_{d,final}$ are the initial and final desired diversities, and all three of them are user-specified algorithmic parameters. For the same reason as the one considered in the three-phased diversity control strategy, we require that the diversity should decrease no slower than linearly with the iteration number. That is, an upper bound $D_{u,n}$ should be set for the diversity, where $D_{u,n}$ is given by

$$D_{u,n} = \frac{(n_{\max} - n)}{(n_{\max})}(D_{u,initial} - D_{u,final}) + D_{u,final}. \tag{58}$$

In (48), $D_{u,initial}$ and $D_{u,final}$ are the initial and final upper bounds for the diversity.

In order to control the diversity of $X_n$ to decrease no faster than $D_{d,n}$ but no slower than $D_{u,n}$, the following strategies are employed. During the evolving process, if $D(X_n)$ is smaller than $D_{d,n}$, the CE coefficient $\alpha$ was set to be $\alpha = \alpha_1 > e^\gamma \approx 1.781$ so that the particles' current positions explode tentatively in order to increase $D(X_n)$ until it is above $D_{d,n}$; if $D(X_n)$ is larger than $D_{u,n}$, the *pbest* position of each particle $P_{i,n}$ is set to be its local focus $p_{i,n}$ in order to accelerate the decline of $D(P_n)$ and, in turn, $D(X_n)$, until $D(X_n)$ is below $D_{u,n}$. This acceleration technique is identical to the one used in the three-phased diversity control strategy when the decline of the diversities is too slow in phase 1. The QPSO with the controlled declining speed of the diversity (QPSO-CDSD) is outlined in Algorithm 3.

**Algorithm 3: The QPSO-CDSD Algorithm**

**Begin**
Initialize the current position $X_{i,0}^j$ and the personal best position $P_{i,0}^j$ of each particle, evaluate their fitness values and find the global best position $G_0 = P_{g,0}$;
Pre-assign $\alpha_1 > 1.781$; Pre-set $D_{d,initial}$, $D_{d,final}$, $D_{u,initial}$ and $D_{u,final}$;
**for** $n = 1$ to $n_{\max}$
    Select the value of $\alpha$ properly;
    Compute the mean best position and calculate $D(X_n)$ by using (49);
    Calculate $D_{d,n}$ and $D_{u,n}$ according to (57) and (58), respectively;
    **if** ( $D(X_n) < D_{d,n}$ )
        $\alpha = \alpha_1$;
    **endif**
    **for** (*i*=1 to *M*)
        **for** *j*=1 to *N*
            $\varphi_{i,n}^j = \text{rand1}(\cdot)$;
            $p_{i,n}^j = \varphi_{i,n}^j P_{i,n}^j + (1 - \varphi_{i,n}^j)G_n^j$;
            $u_{i,n}^j = \text{rand2}(\cdot)$;
            **if** ( $\text{rand3}(\cdot) < 0.5$ )
                $X_{i,n+1}^j = p_{i,n}^j + \alpha \,|\, X_{i,n}^j - C_n^j \,|\, \ln(1/u_{i,n+1}^j)$;
            **else**
                $X_{i,n+1}^j = p_{i,n}^j - \alpha \,|\, X_{i,n}^j - C_n^j \,|\, \ln(1/u_{i,n+1}^j)$;
            **end if**
        **end for**
        Evaluate the objective function value $f(X_{i,n})$;
        Update $P_{i,n}$;
        **if** ( $D(X_n) > D_{u,n}$ )
            $P_{i,n} = p_{i,n}$;
        **end if**
        Update $G_n$;
    **end for**
**end for**
**end**

## 6 Experimental Results on Benchmark Functions

To determine whether the proposed diversity controlling methods can be effective in enhancing the search performance of the QPSO algorithm, a performance comparison using the first fourteen benchmark functions from the CEC 2005 benchmark suite (Suganthan et al., 2005) was carried out between the QPSO-TDC, QPSO-CDSD, standard QPSO algorithms, and other variants of PSO, including PSO with inertia weight (PSO-In) (Shi and Eberhart, 1998a, 1998b, 1999), PSO with constriction factor (PSO-Co) (Clerc, 1999; Clerc and Kennedy, 2002), the PSO-In with local best model (PSO-In-Lbest) (Liang, 2006), the standard PSO (i.e. PSO-Co-Lbest) (Bratton and Kennedy, 2007) , Gaussian PSO (Secrest and Lamont, 2003), Gaussian bare bones PSO (GBBPSO) (Kennedy, 2003, 2004), exponential PSO (PSO-E) (Krohling and Coelho, 2006), Lévy PSO (Richer and Blackwell, 2006), comprehensive learning PSO (CLPSO) (Liang, 2006), dynamic multiple swarm PSO (DMS-PSO) (Liang and Suganthan, 2006) and fully-informed particle swarm (FIPS) (Mendes et al., 2004). For each tested benchmark function, the dimensionality is 30. Each algorithm was run 100 times on each problem, using 20 particles to search the global best fitness value. At each run, the particles started in new and randomly-generated positions, which were uniformly distributed within the search bounds. Each run of each algorithm lasted 10000 iterations (i.e. $n_{\max}=10000$), and the best fitness value (objective function value) for each run was recorded.

For the QPSO algorithm, two control methods for the CE coefficient, namely, the fixed $\alpha$ method and the time-varying $\alpha$ method, were used in our experiments. For convenience of distinction, we denote the QPSO with the fixed CE coefficient as QPSO-FC and the one with the time-varying CE coefficient as QPSO-VC. It has been shown in (Sun et al., 2012b) that fixing $\alpha$ at 0.75 and varying $\alpha$ linearly from 1.0 to 0.5 could yield results with good quality in general for both methods, respectively. Therefore, as in the experiments for diversity testing in Section 5, we use these parameter settings for the QPSO-FC and QPSO-VC.

For the QPSO-TDC algorithm, the value of $d_{lower}$ is set to be $10^{-6}$ as recommended by Usrem and Riget et al. (Ursem 2002; Riget and Vesterstrøm, 2002). The value of $d_{upper}$ was set to be 0.2, which is slightly smaller than the initial distance-to-average-point diversities for each problem. $n_{phase\_1}$ was to be $0.9n_{\max}$, namely 9000 iterations. The CE coefficient in phase 2, $\alpha_2$, was set to be 2.0, consistent with the fact that setting $\alpha_2 > e^\gamma \approx 1.781$ can make the particle swarm explode. The CE coefficient in phase 3 was $\alpha_3=0.75$. For $\alpha_1$, the CE coefficient in phase 1, we can use either the fixed value or the time-varying method. Here, we denote the QPSO-TDC algorithms with the two methods as QPSO-TDC-FC and QPSO-TDC-VC, respectively. In order to evaluate whether the diversity control strategies can improve the search performance of the QPSO, the settings of $\alpha_1$ in the two versions of the QPSO-TDC were the same as those of the CE coefficient in the QPSO-FC and QPSO-VC.

Two versions of the QPSO-CDSD with the two control approaches for the CE coefficient $\alpha$, denoted as QPSO-CDSD-FC and QPSO-CDSD-VC, were tested, with the parameter settings for $\alpha$ the same as those in the QPSO-FC and QPSO-VC. The value of $\alpha_1$ was also set to be 2.0. The parameter $r$ in equation (57) for $D_{d,n}$ was set to be 4 and $D_{d,initial}$ was set to be $(1/3)D(X_0)$, where $D(X_0)$ is the diversity of the initial population of particles' current positions. $D_{u,initial}$ in equation (58) for $D_{u,n}$ was set as $D(X_0)$. Both values of $D_{d,final}$ and $D_{u,final}$ were set to be $10^{-8}$. These parameter settings were recommended according to our preliminary experiments.

The parameter configurations of other PSO variants were the same as those recommended by the existing publications. For the PSO-In, the inertia weight linearly decreased from 0.9 to 0.4 in the course of the run and fixed the acceleration coefficients ($c_1$ and $c_2$) at 2.0, as in the empirical study performed by Shi and Eberhart (Shi and Eberhart, 1999). For the PSO-Co, the constriction factor was set to be $\chi=0.7298$, and the acceleration coefficients $c_1=c_2=2.05$, as recommended by Clerc and Kennedy (Clerc 1999; Clerc and Kennedy, 2002). Eberhart and Shi also used these values of the parameters when comparing the performance of PSO-Co with that of PSO-In (Eberhart and Shi, 2000). For the Standard PSO, the LBEST ring topology was used with other parameter settings the same as those in PSO-Co (Bratton and Kennedy, 2007). Except for the population size and the maximum number of iterations, all the other parameter configurations for the Gaussian bare bones PSO (BBPSO), Gaussian PSO, Lévy PSO, PSO-E, FIPS, DMS-PSO and CLPSO were the same as those recommended in the corresponding papers.

Table 2 lists the mean best fitness value and standard deviation out of 100 runs of the QPSO, QPSO-TCD and QPSO-CDSD algorithms on each problem. The statistical results of the unpaired $t$ tests between the two QPSO and the two diversity-controlled QPSO algorithms, with the two controlling methods for the CE coefficient, are presented in Table 3 and Table 4. Between the QPSO-FC and QPSO-TCD-FC, the latter outperformed significantly the former for four benchmark problems ($F_8$, $F_9$, $F_{10}$ and $F_{13}$). For $F_1$ and $F_2$, the QPSO-TCD-FC had significantly worse performance than the QPSO-FC, since maintaining the diversity above a certain level during the whole search could weaken the local search ability of the QPSO-TCD-FC at the later stage of the search process. For other problems, the two algorithms showed no significant difference. The performance comparison between the QPSO-FC and QPSO-CDSD-FC shows that the QPSO-CDSD-FC generated better results than the QPSO-FC for 10 benchmark functions and the outperformance of the QPSO-CDSD-FC was statistically significant for $F_8$, $F_9$, $F_{10}$, $F_{11}$, $F_{13}$ and $F_{14}$. However, for $F_1$, $F_2$, $F_3$ and $F_6$, the results yielded by the QPSO-CDSD-FC are significantly worse than those yielded by the QPSO-FC. The results also show that the QPSO-TCD-VC had better performance than the QPSO-VC for 8 functions, and for 6 of them, the outperformance of the QPSO-TCD-VC is significant. Except for $F_1$ and $F_4$, the QPSO-CDSD-VC outperformed the QPSO-VC for the other 12 benchmark functions, and for 8 of them, the superiority of the QPSO-CDSD-VC is statistically significant. Generally speaking, the proposed two diversity control strategy, particularly the strategy of controlling the declining speed of the diversity, can improve the search performance of the QPSO for most of the multimodal functions.

Table 2: Mean and Standard Deviation of the Best Fitness Values over 100 Runs of the different versions of QPSO, QPSO-TCD and QPSO-CDSD algorithms

| Algorithms | $F_1$ | $F_2$ | $F_3$ | $F_4$ | $F_5$ | $F_6$ | $F_7$ |
|---|---|---|---|---|---|---|---|
| QPSO-FC | 1.3675e-027 (2.7790e-028) | **1.6945e-012** (**2.5348e-012**) | **6.0409e+005** (**2.8035e+005**) | 210.3383 (712.0292) | 6.7159e+003 (1.8791e+003) | 33.5149 (59.0575) | 0.0183 (0.0172) |
| QPSO-TCD-FC | 5.6179e-011 (4.0244e-011) | 4.9742e-006 (1.3858e-006) | 6.1884e+005 (3.1844e+005) | 321.1697 (569.0590) | 6.7146e+003 (1.5101e+003) | **29.8091** (**59.7588**) | 0.0184 (0.0144) |
| QPSO-CDSD-FC | 4.6031e-012 (1.8094e-012) | 1.6759e-004 (8.0052e-005) | 9.6106e+005 (5.4569e+005) | **90.3147** (**77.2956**) | 3.4048e+003 (959.4392) | 53.8121 (48.2707) | 0.0145 (0.0111) |
| Algorithms | $F_8$ | $F_9$ | $F_{10}$ | $F_{11}$ | $F_{12}$ | $F_{13}$ | $F_{14}$ |
| QPSO-FC | 20.9644 (0.0461) | 40.8432 (13.2970) | 96.3273 (39.1823) | 15.5516 (2.6623) | 4.0545e+003 (4.8607e+003) | 3.8777 (1.2923) | 12.3423 (0.4119) |
| QPSO-TCD-FC | 20.3311 (0.1687) | **8.0518** (**5.0965**) | 74.8887 (23.9318) | 15.2313 (2.5170) | 4.4874e+003 (5.9694e+003) | 3.1859 (1.9102) | 12.3180 (0.4494) |
| QPSO-CDSD-FC | 20.6007 (0.1897) | 11.3425 (7.7929) | 54.8221 (15.2930) | **12.6042** (**2.8866**) | **3.6035e+003** (**3.0482e+003**) | **2.2242** (**0.4811**) | **11.5791** (**0.6844**) |
| Algorithms | $F_1$ | $F_2$ | $F_3$ | $F_4$ | $F_5$ | $F_6$ | $F_7$ |
| QPSO-VC | **7.4132e-028** (**1.7475e-028**) | 0.0293 (0.0259) | 2.5609e+006 (1.3245e+006) | 383.3831 (368.3292) | 3.0126e+003 (1.0897e+003) | 46.8735 (47.6631) | 0.0155 (0.0118) |
| QPSO-TCD-VC | 4.6960e-007 (2.6902e-007) | 0.0478 (0.0605) | 1.6315e+006 (7.0662e+005) | 440.7321 (285.1594) | **2.6754e+003** (**933.8617**) | 69.3284 (59.7424) | **0.0134** (**0.0117**) |
| QPSO-CDSD-VC | 1.2735e-012 (6.6274e-013) | 4.7717e-004 (2.7724e-004) | 1.3057e+006 (6.6584e+005) | 442.7056 (379.0473) | 2.8427e+003 (836.6438) | 43.4228 (31.6621) | 0.0148 (0.0134) |
| Algorithms | $F_8$ | $F_9$ | $F_{10}$ | $F_{11}$ | $F_{12}$ | $F_{13}$ | $F_{14}$ |
| QPSO-VC | 20.9541 (0.0671) | 25.9826 (7.6711) | 80.4498 (44.5904) | 23.9147 (7.2831) | 5.2507e+003 (5.1429e+003) | 3.9523 (1.6955) | 12.5244 (0.5629) |
| QPSO-TCD-VC | **20.3731** (**0.2422**) | 18.0738 (8.0643) | 44.9198 (30.6356) | 20.0522 (5.3951) | 5.8562e+003 (5.0307e+003) | 2.5061 (1.3836) | 12.5271 (0.4826) |
| QPSO-CDSD-VC | 20.6700 (0.1667) | 14.0110 (6.9106) | **43.8545** (**4.6873**) | 14.9255 (3.1176) | 4.4503e+003 (3.9009e+003) | 2.4426 (0.5877) | 11.6102 (0.8167) |

Table 3: T Value and P Value of the Unpaired T Test between the QPSO-FC and QPSO-TCD-FC or QPSO-CDSD-FC

| QPSO-TCD-FC v.s. QPSO-FC | | | | | | | |
|---|---|---|---|---|---|---|---|
| | $F_1$ | $F_2$ | $F_3$ | $F_4$ | $F_5$ | $F_6$ | $F_7$ |
| Stard. Err. | 4.0244e-012 | 1.3858e-007 | 4.2426e-04 | 91.1490 | 241.0688 | 8.4017 | 0.0022 |
| t value | 13.9596 | 35.8941 | 0.3477 | 1.2159 | 0.0054 | 0.4411 | 0.0446 |
| p value | <0.0001 | <0.0001 | 0.7285 | 0.2255 | 0.9957 | 0.6596 | 0.9645 |
| | $F_8$ | $F_9$ | $F_{10}$ | $F_{11}$ | $F_{12}$ | $F_{13}$ | $F_{14}$ |
| Stard. Err. | 0.0175 | 1.4240 | 4.5913 | 0.3664 | 769.8061 | 0.2306 | 0.0610 |
| t value | 36.2123 | 23.0273 | 4.6694 | 0.8742 | 0.5623 | 2.9996 | 0.3986 |
| p value | <0.0001 | <0.0001 | <0.0001 | 0.3830 | 0.5745 | 0.0031 | 0.6906 |
| QPSO-CDSD-FC v.s. QPSO-FC | | | | | | | |
| | $F_1$ | $F_2$ | $F_3$ | $F_4$ | $F_5$ | $F_6$ | $F_7$ |
| Stard. Err. | 1.8094e-013 | 8.0052e-006 | 6.1349e+004 | 71.6212 | 210.9867 | 7.6275 | 0.0020 |
| t value | 25.4399 | 20.9351 | 5.8186 | 1.6758 | 15.6934 | 2.6611 | 1.8563 |
| p value | <0.0001 | <0.0001 | <0.0001 | 0.0954 | <0.0001 | 0.0084 | 0.0649 |

|  | $F_8$ | $F_9$ | $F_{10}$ | $F_{11}$ | $F_{12}$ | $F_{13}$ | $F_{14}$ |
|---|---|---|---|---|---|---|---|
| Stard. Err. | 0.0195 | 1.5412 | 4.2061 | 0.3927 | 573.7415 | 0.2306 | 0.0799 |
| t value | 18.6302 | 19.1410 | 9.8679 | 7.5057 | 0.7861 | 11.9910 | 9.5545 |
| p value | <0.0001 | <0.0001 | <0.0001 | <0.0001 | 0.4328 | <0.0001 | <0.0001 |

Table 4: T Value and P Value of Unpaired T Test between the QPSO-VC and QPSO-TCD-VC or QPSO-CDSD-VC

| QPSO-TCD-VC v.s. QPSO-VC | | | | | | | |
|---|---|---|---|---|---|---|---|
|  | $F_1$ | $F_2$ | $F_3$ | $F_4$ | $F_5$ | $F_6$ | $F_7$ |
| Stard. Err. | 2.6902e-008 | 0.0066 | 1.5012e+005 | 46.5814 | 143.5111 | 7.6426 | 0.0017 |
| t value | 17.4560 | 2.8111 | 6.1910 | 1.2312 | 2.3496 | 2.9381 | 1.2638 |
| p value | <0.0001 | 0.0054 | <0.0001 | 0.2197 | 0.0198 | 0.0037 | 0.2078 |
|  | $F_8$ | $F_9$ | $F_{10}$ | $F_{11}$ | $F_{12}$ | $F_{13}$ | $F_{14}$ |
| Stard. Err. | 0.0251 | 1.1130 | 5.4100 | 0.9064 | 719.4259 | 0.2188 | 0.0741 |
| t value | 23.1177 | 7.1058 | 6.5674 | 4.2615 | 0.8416 | 6.6085 | 0.0364 |
| p value | <0.0001 | <0.0001 | <0.0001 | <0.0001 | 0.4010 | <0.0001 | 0.9710 |
| QPSO-CDSD-VC v.s. QPSO-VC | | | | | | | |
|  | $F_1$ | $F_2$ | $F_3$ | $F_4$ | $F_5$ | $F_6$ | $F_7$ |
| Stard. Err. | 6.6274e-014 | 0.0026 | 1.4824e+005 | 52.8529 | 137.3834 | 5.7221 | 0.0018 |
| t value | 19.2157 | 11.1279 | 8.4671 | 1.1224 | 1.2367 | 0.6030 | 0.3920 |
| p value | <0.0001 | <0.0001 | <0.0001 | 0.2630 | 0.2177 | 0.5472 | 0.6954 |
|  | $F_8$ | $F_9$ | $F_{10}$ | $F_{11}$ | $F_{12}$ | $F_{13}$ | $F_{14}$ |
| Stard. Err. | 0.0180 | 1.0325 | 4.6873 | 0.7922 | 645.4955 | 0.1794 | 0.0992 |
| t value | 15.8099 | 11.5950 | 7.8074 | 11.3467 | 1.2400 | 8.4131 | 9.2167 |
| p value | <0.0001 | <0.0001 | <0.0001 | <0.0001 | 0.2165 | <0.0001 | <0.0001 |

Table 5 and Table 6 record the mean and the standard deviation of the best fitness values out of 100 runs of each algorithm on each problem. To investigate if the differences in mean best fitness values between algorithms were significant, the mean values for each problem were analyzed using a multiple comparison procedure, an ANOVA (Analysis Of Variance) test with 0.05 as the level of significance. The procedure employed in this work is known as the "stepdown" procedure (Day and Quinn, 1989). It was used to determine the algorithmic performance, by ranking for each problem in a statistical manner. The algorithms that were not statistically different to each other were given the same rank; those that were not statistically different to more than one other groups of algorithms were ranked with the best-performing of these groups. For each algorithm, the resulting rank for each problem, the average rank and the final rank are shown in Table 7.

Table 5: Mean and Standard Deviation of the Best Fitness Values after 100 runs of Each Algorithm for $F_1$ to $F_6$

| Algorithms | $F_1$ | $F_2$ | $F_3$ | $F_4$ | $F_5$ | $F_6$ | $F_7$ |
|---|---|---|---|---|---|---|---|
| PSO-In | 2.6399e-25 (1.8131e-24) | 226.8459 (1.0001e+03) | 2.2712e+07 (1.9356e+07) | 2.0077e+03 (1.4855e+03) | 4.7916e+03 (1.6151e+03) | 158.7777 (439.6899) | 0.1027 (0.2033) |
| PSO-Co | 1.0681e-027 (2.1949e-027) | 8.3314e-017 (2.4651e-016) | 7.9934e+006 (9.5976e+006) | 5.6003e+003 (3.5573e+003) | 7.4286e+003 (2.3936e+03) | 54.5151 (63.7504) | 0.0210 (0.0147) |
| PSO-In-Lbest | 2.1805e-27 (1.3803e-26) | 22.4605 (16.4455) | 2.3174e+07 (1.0325e+07) | 6.6424e+03 (2.0921e+03) | 4.8477e+03 (890.8115) | 49.4776 (79.4415) | 0.0178 (0.0276) |
| SPSO (PSO-Co-Lbest) | 1.3127e-29 (5.2814e-29) | 8.5674e-05 (1.1777e-04) | 2.2055e+06 (7.2582e+05) | 4.7982e+03 (2.9767e+03) | 5.7322e+03 (1.5118e+03) | 67.6414 (131.1797) | 0.0185 (0.0139) |
| GBBPSO | 1.3286e-25 (6.8428e-25) | 2.1623e-07 (5.1095e-07) | 3.6361e+06 (1.9245e+06) | 1.1803e+03 (1.2983e+03) | 9.8301e+03 (2.7940e+03) | 53.9102 (103.5611) | 0.0258 (0.0207) |
| Gaussian PSO | 5.7112e-025 (3.9959e-024) | **8.0868e-17** **(1.9412e-16)** | 5.3878e+006 (7.6365e+006) | 1.5014e+004 (8.9493e+003) | 7.0029e+03 (2.1226e+03) | 74.9910 (234.8297) | 0.0244 (0.0220) |
| PSO-E | 1.5368e-025 (4.3049e-025) | 189.6606 (126.3532) | 5.9151e+006 (3.3009e+006) | 3.5301e+003 (1.4726e+003) | 6.8863e+003 (1.5999e+03) | 109.9958 (411.7194) | 0.0263 (0.0372) |
| Levy PSO | 7.8248e-027 (5.0306e-026) | 0.3227 (2.0515) | 2.1908e+07 (2.0804e+07) | 1.4248e+03 (1.3839e+03) | 6.9030e+03 (1.7096e+03) | 69.0370 (131.3992) | 0.1148 (0.2521) |
| FIPS | **4.2495e-032** **(9.3852e-033** | 0.0078 (0.0480) | 2.7516e+06 (2.7027e+06) | 6.9344e+03 (2.4500e+03) | 4.2647e+03 (945.2498) | 71.7734 (121.7390) | 0.0179 (0.0145) |
| DMS-PSO | 1.6715e-020 (7.9818e-020) | 245.3475 (208.9480) | 2.0286e+07 (8.8969e+06) | 4.4459e+03 (1.5628e+03) | 3.2186e+003 (884.6057) | 215.2473 (404.7014) | 0.0220 (0.0173) |
| CLPSO | 1.3796e-031 (2.3856e-032) | 162.4186 (56.9861) | 2.6239e+07 (7.9894e+06) | 3.5384e+03 (1.0470e+03) | 3.4802e+03 (651.3123) | 45.3391 (36.8417) | 0.0973 (0.0518) |
| QPSO-FC | 1.3675e-027 (2.7790e-028) | 2.5348e-012 (2.5348e-012) | 6.0409e+005 (2.8035e+005) | 210.3383 (712.0292) | 6.7159e+003 (1.8791e+03) | 33.5149 (59.0575) | 0.0183 (0.0172) |
| QPSO-VC | 7.4132e-028 (1.7475e-028) | 0.0293 (0.0259) | 2.5609e+006 (1.3245e+006) | 383.3831 (368.3292) | 3.0126e+003 (1.0897e+03) | 46.8735 (47.6631) | 0.0155 (0.0118) |
| QPSO-TDC-FC | 5.6179e-011 (4.0244e-011) | 4.9742e-006 (1.3858e-006) | **6.1884e+005** **(3.1844e+005)** | 321.1697 (569.0590) | 6.7146e+003 (1.5101e+03) | **29.8091** **(59.7588)** | 0.0184 (0.0144) |

| | | | | | | |
|---|---|---|---|---|---|---|
| QPSO-TDC-VC | 4.6960e-007 (2.6902e-007) | 0.0478 (0.0605) | 1.6315e+006 (7.0662e+005) | 440.7321 (285.1594) | **2.6754e+003 (933.8617)** | 69.3284 (59.7424) | **0.0134 (0.0117)** |
| QPSO-CDSD-FC | 4.6031e-012 (1.8094e-012) | 1.6759e-004 (8.0052e-005) | 9.6106e+005 (5.4569e+005) | **90.3147 (77.2956)** | 3.4048e+003 (959.4392) | 53.8121 (48.2707) | 0.0145 (0.0111) |
| QPSO-CDSD-VC | 1.2735e-012 (6.6274e-013) | 4.7717e-004 (2.7724e-004) | 1.3057e+006 (6.6584e+005) | 442.7056 (379.0473) | 2.8427e+003 (836.6438) | 43.4228 (31.6621) | 0.0148 (0.0134) |

Table 6: Mean and Standard Deviation of the Best Fitness Values after 100 runs of Each Algorithm for $F_7$ to $F_{14}$

| Algorithms | $F_8$ | $F_9$ | $F_{10}$ | $F_{11}$ | $F_{12}$ | $F_{13}$ | $F_{14}$ |
|---|---|---|---|---|---|---|---|
| PSO-In | 21.0812 (0.0812) | 32.4966 (14.7031) | 208.4563 (79.5790) | 38.7149 (7.3471) | 2.7119e+004 (2.7609e+004) | 4.1789 (3.0407) | 13.7076 (0.2690) |
| PSO-Co | 21.0774 (0.0847) | 87.6755 (22.8381) | 127.1855 (48.5208) | 28.8098 (4.2308) | 1.8188e+004 (2.2818e+004) | 5.3719 (2.0890) | 12.8125 (0.5639) |
| PSO-In-Lbest | 20.8930 (0.0604) | 40.0970 (9.5686) | 142.8008 (44.4868) | 29.6431 (2.2897) | 2.0623e+004 (1.1976e+004) | 4.8445 (1.2006) | 12.9876 (0.2571) |
| SPSO (PSO-Co-Lbest) | 20.9093 (0.0574) | 74.7213 (18.6954) | 95.3674 (28.0466) | 30.0476 (2.0667) | 4.9388e+003 (4.5279e+003) | 5.2913 (1.1650) | 12.6469 (0.4580) |
| GBBPSO | 20.9661 (0.0424) | 88.8495 (25.2804) | 143.4723 (44.9368) | 37.7460 (4.1352) | 2.4132e+004 (5.1706e+004) | 5.5438 (1.7552) | 13.2691 (0.5413) |
| Gaussian PSO | 21.0728 (0.0818) | 94.5406 (25.1829) | 130.6175 (39.7167) | 28.6698 (4.1138) | 4.4787e+003 (8.7792e+003) | 6.3620 (1.9880) | 12.9898 (0.4007) |
| PSO-E | 20.9532 (0.0590) | 50.2256 (16.1649) | 141.9179 (56.2953) | 25.3028 (3.2794) | 1.1181e+004 (7.8156e+003) | 3.0675 (0.9151) | 13.0002 (0.3309) |
| Levy PSO | 21.0954 (0.0596) | 49.1907 (15.7793) | 166.1596 (85.9732) | 27.8213 (5.2623) | 2.7182e+004 (3.0088e+004) | 6.6914 (2.9238) | 13.4094 (0.3772) |
| FIPS | 20.9512 (0.0518) | 41.3979 (10.9162) | 167.2160 (31.8707) | 31.3372 (3.5362) | 2.9583e+004 (1.6172e+004) | 9.3897 (1.4136) | 12.8095 (0.2513) |
| DMS-PSO | 20.7905 (0.0938) | 47.1811 (11.1399) | 110.2614 (22.5910) | 28.5874 (1.7909) | 1.8724e+004 (1.6029e+004) | 5.1676 (1.8393) | 12.7630 (0.2660) |
| CLPSO | 20.9723 (0.0542) | **0.0995 (0.3015)** | 90.5630 (14.8545) | 26.7164 (1.9216) | 2.3055e+004 (7.3705e+003) | 2.6973 (0.3055) | 13.0663 (0.2132) |
| QPSO-FC | 20.9644 (0.0461) | 40.8432 (13.2970) | 96.3273 (39.1823) | 15.5516 (2.6623) | 4.0545e+003 (4.8607e+003) | 3.8777 (1.2923) | 12.3423 (0.4119) |
| QPSO-VC | 20.9541 (0.0671) | 25.9826 (7.6711) | 80.4498 (44.5904) | 23.9147 (7.2831) | 5.2507e+003 (5.1429e+003) | 3.9523 (1.6955) | 12.5244 (0.5629) |
| QPSO-TDC-FC | **20.3311 (0.1687)** | 8.0518 (5.0965) | 74.8887 (23.9318) | 15.2313 (2.5170) | 4.4874e+003 (5.9694e+003) | 3.1859 (1.9102) | 12.3180 (0.4494) |
| QPSO-TDC-VC | 20.3731 (0.2422) | 18.0738 (8.0643) | 44.9198 (30.6356) | 20.0522 (5.3951) | 5.8562e+003 (5.0307e+003) | 2.5061 (1.3836) | 12.5271 (0.4826) |
| QPSO-CDSD-FC | 20.6007 (0.1897) | 11.3425 (7.7929) | 54.8221 (15.2930) | **12.6042 (2.8866)** | **3.6035e+003 (3.0482e+003)** | **2.2242 (0.4811)** | **11.5791 (0.6844)** |
| QPSO-CDSD-VC | 20.6700 (0.1667) | 14.0110 (6.9106) | **43.8545 (4.6873)** | 14.9255 (3.1176) | 4.4503e+003 (3.9009e+003) | 2.4426 (0.5877) | 11.6102 (0.8167) |

Table 7: Ranking by Algorithms and Problems Obtained from "Stepdown" Multiple Comparisons

| Algorithms | $F_1$ | $F_2$ | $F_3$ | $F_4$ | $F_5$ | $F_6$ | $F_7$ | $F_8$ | $F_9$ | $F_{10}$ | $F_{11}$ | $F_{12}$ | $F_{13}$ | $F_{14}$ | Ave. Rank | Final Rank |
|---|---|---|---|---|---|---|---|---|---|---|---|---|---|---|---|---|
| PSO-In | =6 | =15 | =13 | 9 | =8 | =16 | =15 | =14 | 7 | 17 | 17 | =10 | 9 | 17 | 12.36 | 17 |
| PSO-Co | 5 | =1 | 12 | 14 | =11 | =7 | =10 | =14 | 15 | =10 | 12 | =10 | 13 | =8 | 10.14 | 13 |
| PSO-In-Lbest | =6 | 13 | =13 | =15 | =8 | =1 | =1 | =6 | 8 | =10 | 13 | =10 | 10 | =11 | 8.93 | 9 |
| SPSO (PSO-Co-Lbest) | 3 | 6 | 6 | =12 | 10 | =7 | =1 | =6 | 14 | =6 | 14 | =1 | 12 | =5 | 7.36 | 7 |
| GBBPSO | =6 | 4 | 9 | =7 | 17 | =7 | =10 | =8 | 16 | =10 | 16 | 10 | 14 | 15 | 10.64 | 15 |
| Gaussian PSO | =6 | =1 | =10 | 17 | =11 | =7 | =10 | =14 | 17 | =10 | 11 | =1 | 15 | =11 | 10.07 | 12 |
| PSO-E | =6 | =15 | =10 | =10 | =11 | =7 | =10 | =8 | 13 | =10 | 7 | 9 | 5 | =11 | 9.43 | 10 |
| Levy PSO | =6 | =11 | =13 | =7 | =11 | =7 | =15 | =14 | 12 | =15 | 9 | =10 | 16 | 16 | 11.57 | 16 |
| FIPS | 1 | =8 | =7 | =15 | 7 | =7 | =1 | =8 | 10 | =15 | 15 | 17 | 17 | =8 | 9.71 | 11 |
| DMS-PSO | 13 | =15 | =13 | =12 | =3 | =16 | =10 | 5 | 11 | 9 | 10 | =10 | 11 | =8 | 10.43 | 14 |
| CLPSO | 2 | 14 | 17 | =10 | =5 | =1 | =15 | =8 | 1 | =6 | 8 | =10 | =2 | =11 | 7.86 | 8 |
| QPSO-FC | =6 | 3 | =1 | =1 | =1 | =1 | =1 | =8 | 9 | =6 | 4 | =1 | 7 | =3 | 4.43 | 4 |
| QPSO-VC | 4 | 10 | =7 | =3 | =3 | =1 | =1 | =8 | 6 | =4 | 6 | =1 | 8 | =5 | 4.79 | 6 |
| QPSO-TDC-FC | 16 | 5 | =1 | =3 | =11 | =1 | =1 | =1 | 2 | =4 | 3 | =1 | 6 | =3 | 4.14 | 3 |
| QPSO-TDC-VC | 17 | =11 | 5 | =3 | =1 | =7 | =1 | =1 | 5 | =1 | 5 | =1 | =2 | =5 | 4.64 | 5 |
| QPSO-CDSD-FC | 15 | 7 | 3 | =1 | =5 | =7 | =1 | 3 | 3 | 3 | 1 | =1 | 1 | =1 | 3.71 | 2 |
| QPSO-CDSD-VC | 14 | =8 | 4 | =3 | =1 | =1 | =1 | 4 | 4 | =1 | 2 | =1 | =2 | =1 | 3.36 | 1 |

For the Shifted Sphere Function ($F_1$), the FIPS generated better results than other methods. The QPSO-TDC and QPSO-CDSD showed inferior performance to the other methods since diversity control strategies may weaken the local search ability of the algorithm. The results for the Shifted Schwefel's Problem 1.2 ($F_2$) show that the PSO-Co and Gaussian PSO yielded the best results, but the performance of the DMS-PSO seems to be inferior to that of other competitors due to its slow convergence speed. For the Shifted Rotated High Conditioned Elliptic Function ($F_3$), the QPSO-TDC-FC and QPSO-FC outperformed the other methods in a statistical significance manner. The QPSO-CDSD-FC and QPSO-FC algorithms were showed to be the winner among all the tested algorithms for the Shifted Schwefel's Problem 1.2 with Noise in Fitness ($F_4$). $F_5$ is the Schwefel's Problem 2.6 with Global Optimum on the Bounds. For this benchmark, the QPSO-TDC-VC shared the first place with the QPSO-CDSD-VC from the perspective of the statistical test. For benchmark $F_6$, the Shifted Rosenbrock Function, all the QPSO variants except the QPSO-CDSD-FC and QPSO-CDSD-FC, CLPSO and PSO-In were tied for the first place. The results for the Shifted Rotated Griewank's Function without Bounds ($F_7$) suggest that all the QPSO variants, the standard PSO, the PSO-In-Lbest and the FIPS were able to find the solution to the function with better quality compared to the other methods. Benchmark $F_8$ is the Shifted Rotated Ackley's Function with Global Optimum on the Bounds. Both versions of the QPSO-TDC yielded better results for this problem than the others. The Shifted Rastrigin's Function ($F_9$) is a separable function, which the CLPSO algorithm was good at solving it, and obtained remarkably better results. It can also be observed that the QPSO-TDC-FC yielded a better result than the remainders. $F_{10}$ is the Shifted Rotated Rastrigrin's Function, which appears to be a more difficult problem than $F_9$. For this benchmark, both the QPSO-CDSD-VC and QPSO-TDC-VC outperformed the other competitors in a statistically significant manner. The best results for the Shifted Rotated Weierstrass Function ($F_{11}$) were obtained by the QPSO-CDSD-FC. When searching the optima of Schwefel's Problem 2.13 ($F_{12}$), all the QPSO variants, the Gaussian and the SPSO were found to tie for first rank in a statistical manner. $F_{13}$ is the Shifted Expand Griewank's plus Rosenbrock's Function, for which the QPSO-CDSD-FC obtained better results than its competitors. The ranks of QPSO-CDSD-VC, QPSO-TDC-VC and CLPSO shared the second for this function. For Shifted Rotated Expanded Scaffer's $F6$ Function ($F_{14}$), the two versions of the QPSO-CDSD showed statistically better performance than the others.

The average ranks and final ranks listed in Table 7 reveal that the QPSO-CDSD-VC had the best overall performance among all the tested algorithms. Except for $F_1$ and $F_2$, it had fairly stable performance across all the other tested benchmark functions, with the worst rank being 4 for $F_3$, $F_7$ and $F_8$, respectively. The second best-performing is the QPSO-CDSD-FC. For six of the fourteen benchmark functions, the algorithm had the first performance rank, but unsatisfactory performance for $F_1$ and $F_2$, as the QPSO-CDSD-VC did, due to weak exploitation ability resulted from the diversity maintenance at the later stage of the search process. The next best performing algorithm was the QPSO-TDC-FC with the average rank being 4.14. The overall performance of the QPSO-TDC-VC is no better than that of the QPSO-FC, but it showed some advantages over the QPSO-VC, which employs the same controlling method for the CE coefficient. However, to sum up, the three-phased diversity controlling strategy indeed enhanced the search performance of the QPSO, as can be found from the total average rank of the two QPSO-TDC algorithms (equal to 4.39) and that of the two QPSO algorithms (equal to 4.61). Between the two diversity controlling strategies, controlling the declining speed of the diversity seemed to be more effective in enhancing the search performance of the QPSO algorithm. As for the controlling method for the CE coefficient, it is preferable to use fixed $\alpha$ for the QPSO-TDC algorithm and to employ time-varying $\alpha$ for the QPSO-CDSD algorithm. It can be seen from the average ranks of the QPSO-FC and QPSO-VC that for the original QPSO, employing fixed $\alpha$ was more supportive for the algorithmic performance when the population size and the maximum number of iterations are given.

Except for the QPSO and its variants with the diversity controlling strategies, the best-performing algorithm was the PSO-Co-Lbest, i.e., the standard PSO. For $F_7$ and $F_{12}$, it yielded the results sharing the best place. The next best algorithm is the CLPSO, which is very effective in solving separable functions such as $F_9$, but not some unimodal ones, such as $F_2$ and $F_3$. The PSO-In-Lbest is the PSO-In with the ring neighborhood topology and it obtained better overall performance than the PSO-In and the remainder competitors. For $F_6$ and $F_7$, its performance ranks shared the first place with others, which may be mainly due to the ring topology. Therefore, it is conclusive from Table 6 that incorporating the ring topology into the PSO-In and PSO-Co could enhance the overall performances of the two PSO variants on the tested benchmark functions. The FIPS, which also used the ring topology, found the best results for $F_1$ and $F_7$. Among the probabilistic PSO variants, the PSO-E was the best performing one. The Gaussian PSO and GBBPSO were showed to have good performance for $F_2$, which implies that they had relatively faster convergence for this benchmark function.

## 7 Conclusions

In this paper, we analyzed the correlation between the diversities and the search performance, and proposed two diversity controlling strategies for the QPSO, which were subsequently tested on a set of benchmark functions. The genotype and phenotype diversities of the QPSO were measured by the distance to the average point and the proportional entropy of the fitness values, respectively. Correlations between the diversities and the search performance of the algorithm were tested and analyzed on several widely used benchmark functions. We found that low distance-to-average-point diversities always accompany good fitness values, and there is a strong association between the changes in the improvement of the best fitness with the trends of the correlation coefficients between the distance-to-average-point diversities and the best fitness. Therefore, it was concluded that the distance-to-average-point diversity plays a more important role in the evolving process of the QPSO than the entropy diversities.

Based on the analysis of correlations between diversities and the best fitness, two strategies were proposed to control the distance-to-average-point diversities in QPSO in order to improve the algorithmic performance. The improved QPSO with the proposed diversity controlling approaches, along with the original QPSO and other PSO variants, were tested on the first fourteen benchmark functions from the CEC 2005 benchmark suite. The performance comparisons among the tested algorithms showed that the proposed diversity controlling strategies were effective in enhancing the search ability of the QPSO algorithm.

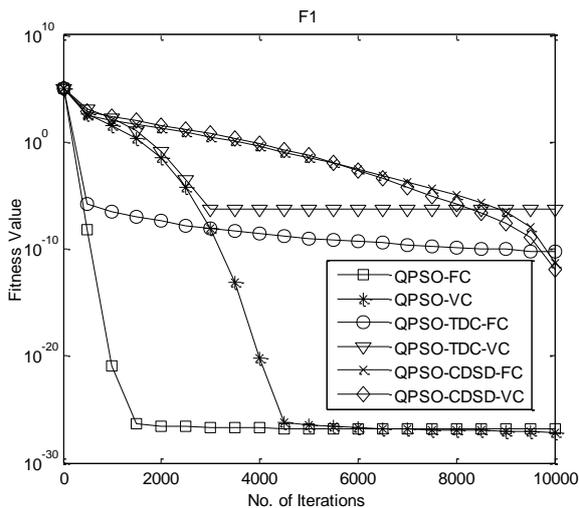
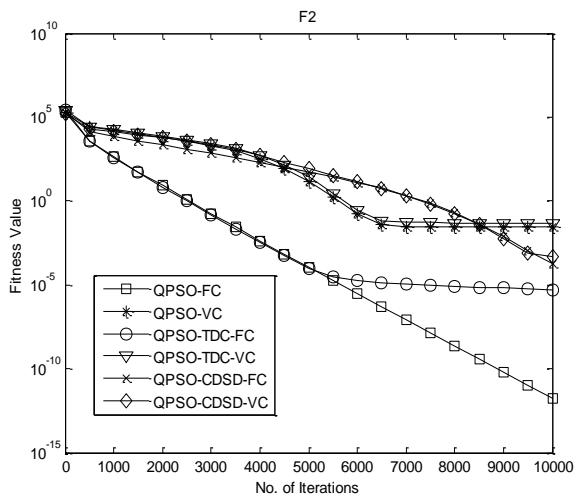
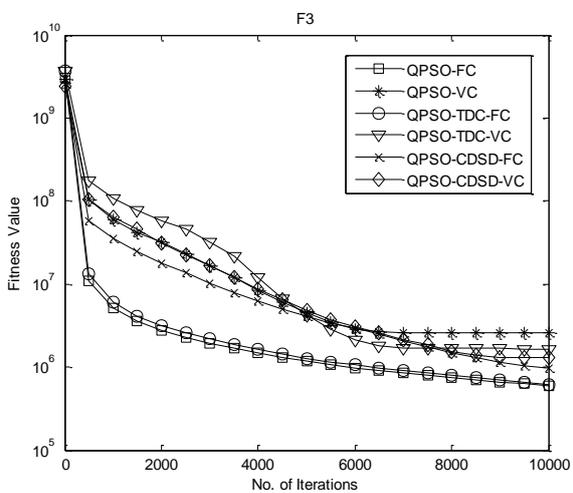
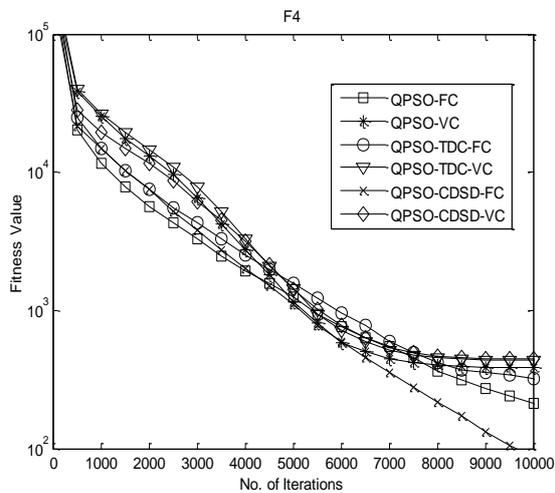
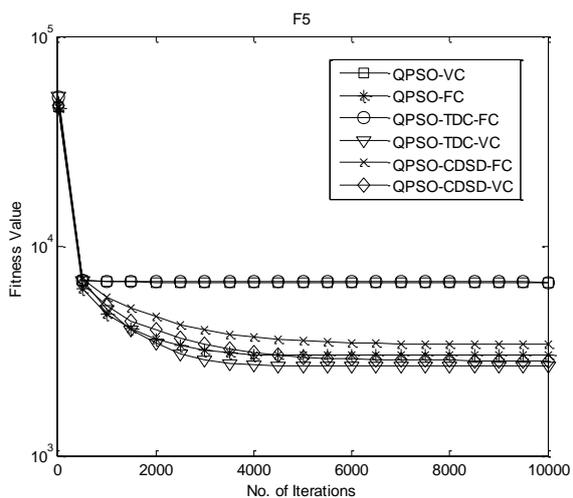
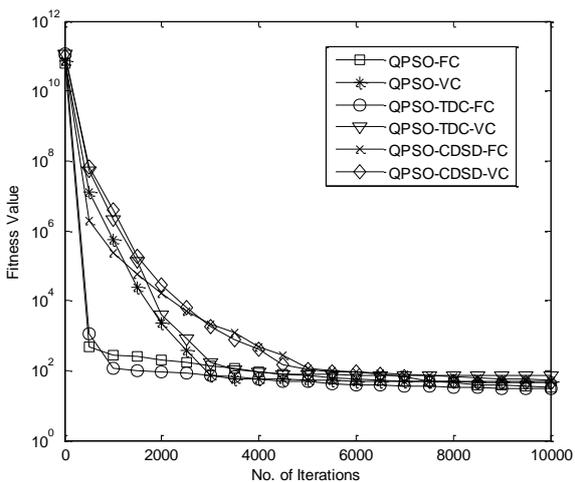

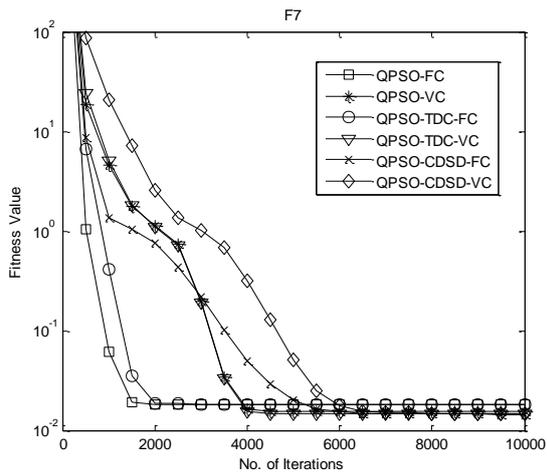
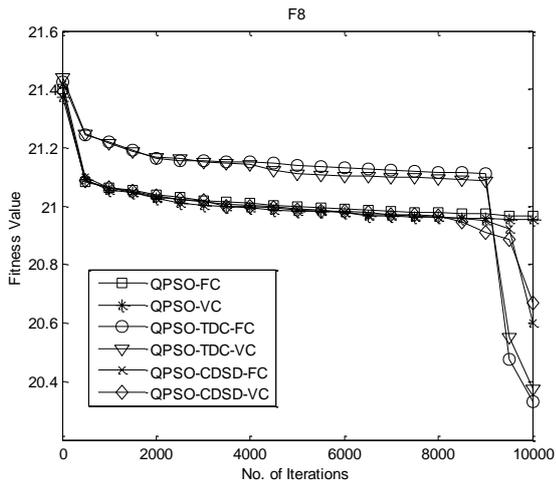
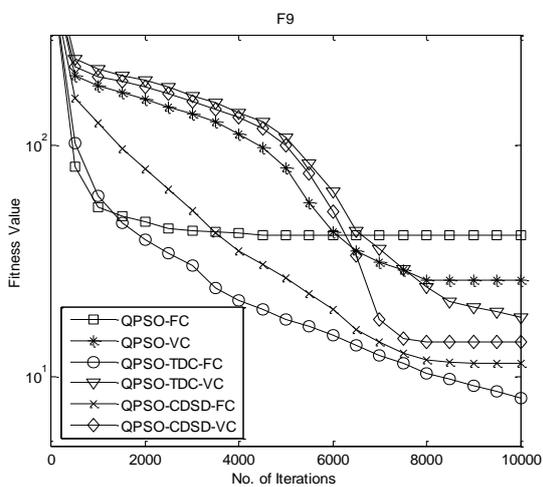
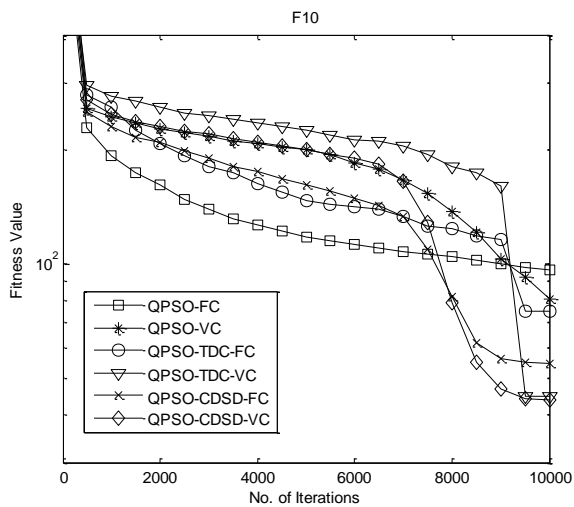
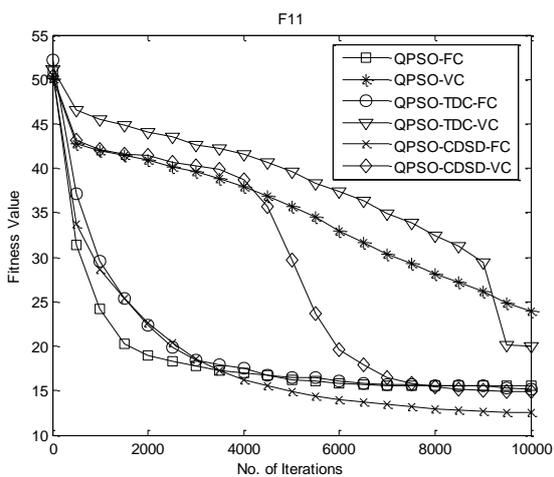
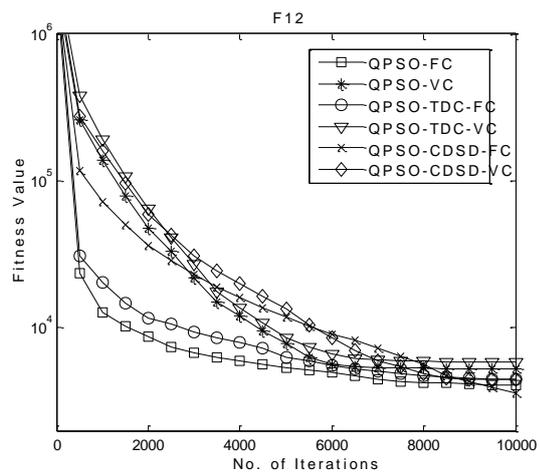

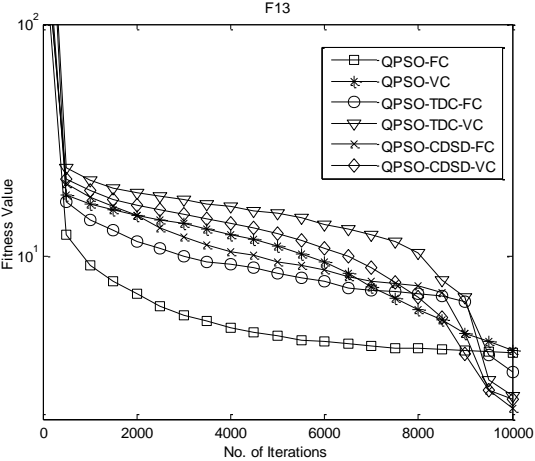 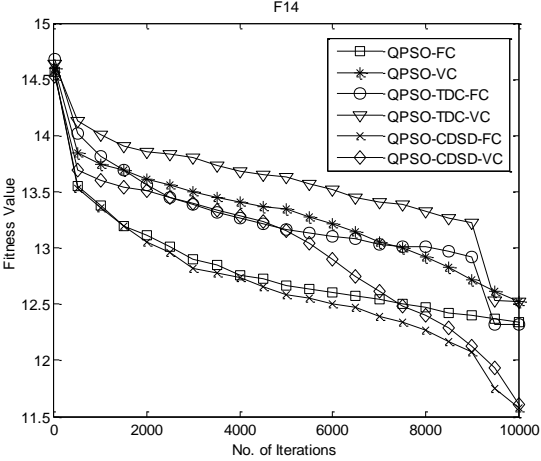